\documentclass[runningheads]{llncs}

\usepackage{eccv}

\usepackage{eccvabbrv}

\usepackage{graphicx}
\usepackage{booktabs}

\usepackage[accsupp]{axessibility}  

\usepackage{orcidlink}
\definecolor{cvprblue}{rgb}{0.21,0.49,0.74}
\definecolor{ibmred}{rgb}{0.905,0.435,0.317}
\definecolor{mygray}{rgb}{.929, .929, .913}
\definecolor{mygreen}{RGB}{0,128,0}
\definecolor{myred}{RGB}{255,0,0}
\usepackage{multirow}
\usepackage{pifont}
\usepackage{arydshln}
\usepackage{colortbl}
\usepackage{xcolor}
\usepackage{ulem} 
\usepackage{comment}
\usepackage{tablefootnote}
\usepackage{threeparttable}
\usepackage{subcaption}
\usepackage{longtable}
\usepackage{tabularx}
\usepackage{multicol}
\usepackage{wrapfig}

\begin{document}

\title{From One-to-One to Many-to-Many: Dynamic Cross-Layer Injection for Deep Vision-Language Fusion} 

\titlerunning{CLI: Dynamic Cross-Layer Injection for Deep Vision-Language Fusion}

\author{Cheng Chen\inst{1,2}\textsuperscript{$\ast$} \and
Yuyu Guo\inst{2}\textsuperscript{$\ast$} \and
Pengpeng Zeng\inst{1} \and
Jingkuan Song\inst{1,3}\textsuperscript{$\dag$} \and
Peng Di\inst{2} \and
Hang Yu\inst{2}\textsuperscript{$\dag$} \and
Lianli Gao \inst{4}}

\authorrunning{C. Chen et al.}

\institute{Tongji University, China \and
Ant Group, China \and
Shanghai Innovation Institute, China \and 
Independent Researcher, China\\
\email{\{cczacks,yuyuguo1994,jingkuan.song\}@gmail.com}, hyu1@e.ntu.edu.sg}

\footnotetext[1]{$\ast$ Equal contribution.}
\footnotetext[2]{$\dag$ Corresponding author.}

\maketitle

\begin{abstract}
Vision-Language Models (VLMs) create a severe visual feature bottleneck by using a crude, asymmetric connection that links only the output of the vision encoder to the input of the large language model (LLM). 
This static architecture fundamentally limits the ability of LLMs to achieve comprehensive alignment with hierarchical visual knowledge, compromising their capacity to accurately integrate local details with global semantics into coherent reasoning. \footnote{Note that \textit{reasoning} throughout this paper denotes the model's general inferential ability to interpret and integrate visual and linguistic cues, rather than any explicit ``thinking'' paradigm.}
To resolve this, we introduce \textbf{Cross-Layer Injection (CLI)}, a novel framework that forges a dynamic ``\textbf{many-to-many}'' bridge between the two modalities. 
CLI consists of two synergistic, parameter-efficient components: an \textbf{Adaptive Multi-Projection (AMP)} module that harmonizes features from diverse vision layers, and an \textbf{Adaptive Gating Fusion (AGF)} mechanism that empowers the LLM to selectively inject the most relevant visual information based on its real-time decoding context. 
We validate the effectiveness and versatility of CLI by integrating it into LLaVA-OneVision and LLaVA-1.5. 
Extensive experiments on 28 diverse benchmarks demonstrate significant performance improvements, establishing CLI as a scalable paradigm that unlocks deeper multimodal understanding by granting LLMs on-demand access to the full visual hierarchy. Code is available at \url{https://github.com/codefuse-ai/CLI}.

  \keywords{Vision-Language Model \and Multimodal Large Language Model}
\end{abstract}

\section{Introduction}

Vision-Language Models (VLMs)~\cite{DBLP:conf/nips/AlayracDLMBHLMM22, DBLP:conf/icml/0008LSH23, DBLP:conf/nips/LiuLWL23a, DBLP:conf/cvpr/LiuLLL24} are increasingly viewed as a pivotal step toward Artificial General Intelligence. 
By endowing powerful Large Language Models (LLMs) with sight from Vision Transformers (ViTs)~\cite{DBLP:conf/icml/RadfordKHRGASAM21,DBLP:conf/iccv/ZhaiM0B23}, VLMs can perceive and reason about the world in a manner that more closely mirrors human cognition. 
Beneath this ambition, however, lies a profound architectural \textbf{symmetry} between ViTs and LLMs that most current designs shatter. 
Both progressively refine representations~\cite{starace2023probing,song2025demystifying,amir2021deep}, yet they are connected by a crude, \textbf{asymmetric} bridge where only the top of the vision hierarchy is linked to the base of the language hierarchy. 
Such static and single-sided connection breaks the intrinsic hierarchical symmetry between ViTs and LLMs~\cite{starace2023probing,song2025demystifying,amir2021deep}, leading to limited fine-grained perception and shallow multimodal reasoning.

\begin{wrapfigure}{t}{0.5\linewidth}
\centering  
\vspace{-2ex}
\includegraphics[width=1.0\linewidth]{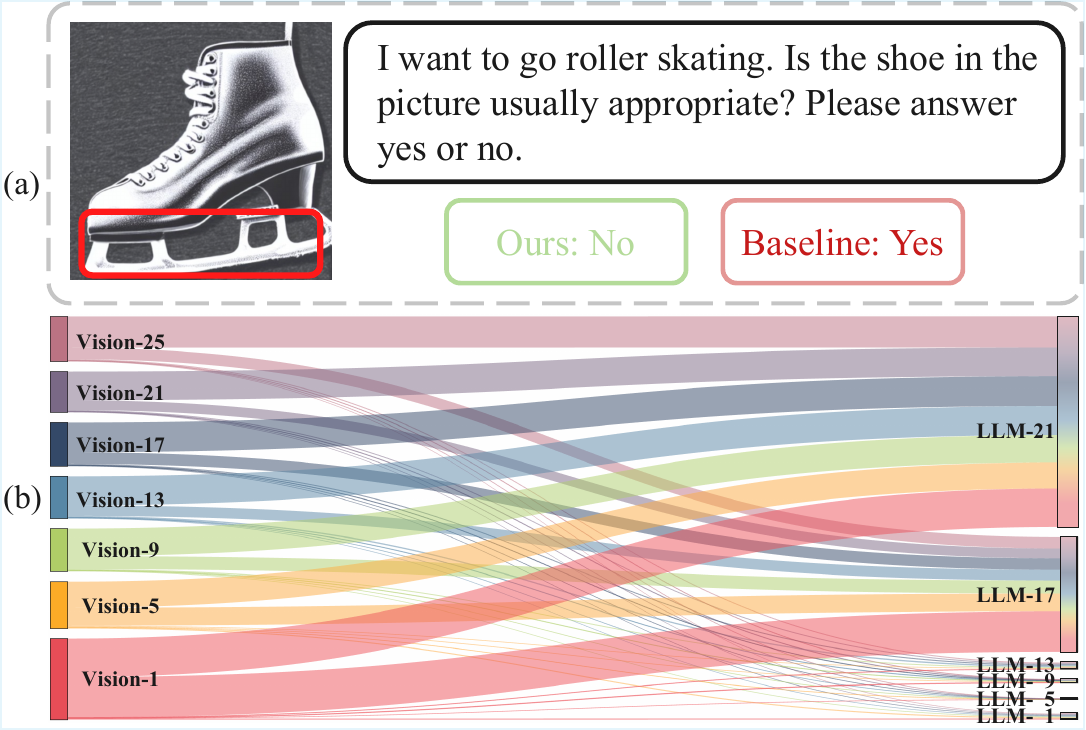}
\caption{(a) An illustrative failure case where our baseline model, relying solely on final-layer features, incorrectly identifies an ice skate as `roller skates'.
This error underscores the necessity of accessing fine-grained details from early vision encoder layers.
(b) Visualized gating weights from our CLI framework validate the efficacy of ``criss-cross connections''. 
The heatmap reveals that deeper LLM decoder layers (right) dynamically query features from the full spectrum of vision encoder layers (left)---a many-to-many interaction that confirms our proposed fusion architecture.
}
\vspace{-2ex}
\label{fig:attention_sparsity}
\end{wrapfigure}

This limitation has tangible consequences, as shown in Fig.~\ref{fig:attention_sparsity} (a).
A baseline VLM, relying only on final-layer features, recognizes the object as ``roller skates'' but overlooks critical wheel details, thus failing to assess its usability.
This failure highlights a fundamental \textbf{functional mismatch}: the static, single-layer connection denies the LLM access to the full spectrum of visual evidence, preventing it from integrating local details with global semantics for nuanced reasoning.

An effective integration must accommodate the diverse and dynamic needs of the LLM processing hierarchy.
The most intuitive requirement is a parallel alignment: shallow LLM layers, processing local syntax~\cite{starace2023probing}, need access to early ViT layers capturing edges and textures~\cite{amir2021deep} to ground basic nouns. 
Similarly, deep LLM layers handling abstract reasoning~\cite{song2025demystifying} would logically leverage high-level scene semantics from deep ViT layers~\cite{amir2021deep}. 
Yet, even this parallel structure is still insufficient. 
The demands of sophisticated multimodal reasoning also necessitate \textbf{criss-crossed connections}. 
For instance, when processing the prompt in Fig.~\ref{fig:attention_sparsity}, a shallow LLM layer parsing the word “shoe” requires immediate access to the high-level, semantic concept of the object—an abstraction formed only in the deep ViT layers—to correctly ground the noun. 
Simultaneously, to answer the ultimate question, “Is the shoe...appropriate?”, a deep LLM layer performing this evaluative reasoning must “zoom in” on the fine-grained structural details of the wheels—information captured only in the early ViT layers. 
This single example powerfully illustrates the need for both types of criss-crossed connections. 
Indeed, our own experiments confirm this necessity, revealing a complex pattern of learned connections (Fig.~\ref{fig:attention_sparsity} (b)) that move far beyond a simple one-to-one mapping.

Solving this requires a fundamental shift from the prevailing one-to-one connections to a truly dynamic many-to-many framework.
However, prior research attempting to bridge this gap offers only incomplete solutions:
approaches like DeepStack~\cite{DBLP:conf/nips/MengYTDW0024} provide a form of brute-force one-to-many injection. 
Other models, from EVLM~\cite{DBLP:journals/corr/abs-2407-14177} to the recent Qwen3-VL~\cite{qwen3vl}, impose a heavyweight and rigid one-to-one fusion, hard-wiring specific vision layers to pre-configured LLM layers. 
All such methods create a static ``straitjacket'', failing to provide the dynamic, arbitrary access that sophisticated reasoning demands.

\begin{wrapfigure}{t}{0.5\linewidth}
\centering  
\vspace{-2ex}
\includegraphics[width=1.0\linewidth]{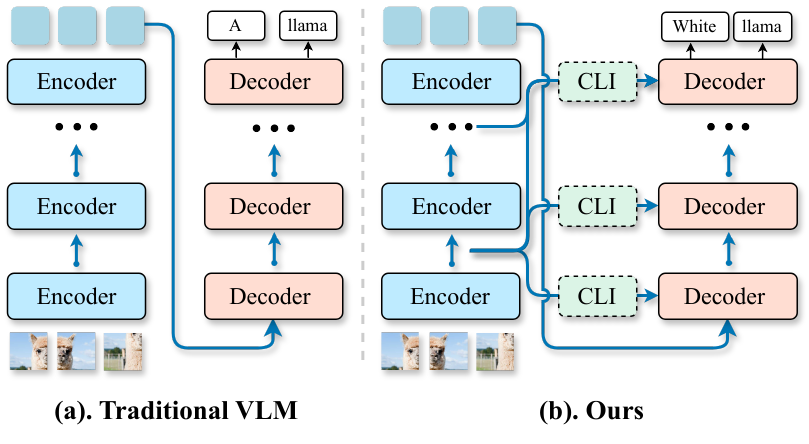}
\caption{(a) Conventional VLM Pipeline. Maps visual features from the encoder's final layer to the text embedding space via a projector.
(b) ``Many-to-Many'' Bridge. Adaptively fuses features from multiple vision layers into multiple LLM layers.
}
\vspace{-2ex}
\label{fig:traditional_mllm}
\end{wrapfigure}
We argue that the solution is to empower the LLM to become an \textbf{active observer}, capable of dynamically querying the entire visual hierarchy at the precise level of detail required by its current reasoning step. 
To this end, we are among the \textbf{first} to propose \textbf{Cross-Layer Injection (CLI)} (as shown in Fig.~\ref{fig:traditional_mllm} (b)), a framework designed to realize an adaptive \textbf{many-to-many} architecture for VLMs. 
CLI consists of two synergistic, parameter-efficient components. 
First, \textbf{Adaptive Multi-Projection} uses Low-Rank Adaptation (LoRA) \cite{DBLP:conf/iclr/HuSWALWWC22} to efficiently harmonize features from diverse vision layers into a shared semantic space. 
Second, and most critically, our \textbf{Adaptive Gating Fusion} mechanism acts as an intelligent, context-sensitive controller. 
At each injection point, it allows the LLM to query this multi-level visual repository and selectively integrate only the most relevant information—be it fine-grained textures from a shallow ViT layer or abstract concepts from a deep one.

To validate the effectiveness and broad applicability of our CLI framework, we integrated it into two distinct VLM architectures: LLaVA-OneVision and LLaVA-1.5. 
Across 28 diverse and challenging benchmarks—covering key capabilities like document analysis, chart reasoning, and general visual perception—our method significantly outperforms the baseline models and other deep fusion strategies. 
In particular, when applied to LLaVA-OV-7B, our method achieves performance improvements of \textbf{6.5}, \textbf{3.3}, and \textbf{4.7} points on the LLaVA-in-the-Wild, MME, and the OCR-Bench benchmarks, respectively.
In summary, our main contributions are as follows:
\begin{itemize} 
\item A novel framework, CLI, that systematically addresses the underutilization of hierarchical visual features in VLMs.
\item Two synergistic and parameter-efficient modules, Adaptive Multi-Projection and Adaptive Gating Fusion, that collectively enable the alignment and dynamic, context-sensitive fusion of multi-level visual features.
\item Extensive experiments on two distinct VLM architectures and a broad range of diverse benchmarks, demonstrating consistent and significant performance gains that establish a new state-of-the-art (SOTA) on several key tasks.
\end{itemize}

\section{Related Works}
\subsection{The Dominant VLM Paradigm}
The paradigm shift in AI driven by LLMs~\cite{DBLP:conf/nips/BrownMRSKDNSSAA20, DBLP:journals/corr/abs-2302-13971} has extended into the visual domain, leading to the rapid emergence of VLMs. 
A dominant architectural blueprint, established by seminal works like Flamingo \cite{DBLP:conf/nips/AlayracDLMBHLMM22} and BLIP-2 \cite{DBLP:conf/icml/0008LSH23} and popularized by LLaVA~\cite{DBLP:conf/nips/LiuLWL23a, DBLP:conf/cvpr/LiuLLL24}, is rooted in parameter-efficient adaptation. This approach involves freezing the powerful pre-trained vision and language backbones and training only a lightweight ``bridge'' module between them. 
This powerful and efficient design crystallized the ``pre-training + fine-tuning'' paradigm that now dominates the field, sparking a wave of open-source innovation, including InstructBLIP \cite{DBLP:conf/nips/Dai0LTZW0FH23}, MiniGPT-4 \cite{DBLP:journals/corr/abs-2310-09478,DBLP:conf/iclr/Zhu0SLE24}, Qwen-VL \cite{DBLP:journals/corr/abs-2308-12966}, and InternVL-2.5 \cite{DBLP:journals/corr/abs-2412-05271}. Beneath this success, however, lies the fundamental architectural flaw we identify in our introduction: these models almost universally treat the output of the final vision encoder layer as the sole representation of the image, creating a severe information bottleneck and limiting the LLM's perceptual depth.

\subsection{Deeper Fusion: Hierarchical Approaches}

To address the above problem, subsequent research has explored deeper visual-language fusion strategies. 

\vspace{1.0ex}
\noindent\textbf{One-to-Many Injection: Brute-Force and Context-Blind.} 
The first category attempts to inject features from a single image (one) into multiple LLM layers (many) through direct, non-adaptive methods. DeepStack~\cite{DBLP:conf/nips/MengYTDW0024} is a prominent example of this approach, which partitions high-resolution visual tokens into distinct groups and feeds each group to a different LLM decoder layer via simple element-wise addition. 
This brute-force mechanism is context-blind; the non-adaptive summation risks disrupting the LLM’s carefully learned representations by injecting unfiltered, and potentially irrelevant, visual information. 
CogVLM \cite{wang2024cogvlm} integrates ``Visual Expert'' modules into each LLM layer, these modules all process the same, single feature map from the vision encoder's final layer.
In addition, a distinct approach is taken by CogAgent \cite{hong2024cogagent}, whose primary goal is high-resolution efficiency.
It employs a parallel architecture with a separate, lightweight encoder for high-resolution details.
These features are statically broadcast to each decoder layer, meaning every layer accesses the same, fixed level of visual information.
Other work, such as FUSION~\cite{DBLP:journals/corr/abs-2504-09925}, deepens the fusion process by introducing ``interaction layers'' where learnable tokens recursively engage with both textual and visual features. 
However, this approach focuses on the interaction and still operates on a single level of visual representation.

\vspace{1.0ex}
\noindent\textbf{Statically-Wired One-to-One Fusion: Rigid and Inflexible.} 
The second category employs dedicated modules to create fixed connections between specific vision and LLM layers, forming a kind of one-to-one mapping at different points in the hierarchy. 
The work of \cite{lin2025multi} systematically compared various internal fusion methods, demonstrating that simple, static approaches often fail.
Flamingo-style architectures like EVLM~\cite{DBLP:journals/corr/abs-2407-14177} exemplify this by inserting a new cross-attention layer before each LLM layer and feeding it features from a specific, pre-assigned ViT layer. 
This philosophy of sparsely inserting specialized fusion blocks into a pre-configured structure is also seen in models like mPLUG-Owl3~\cite{yem2025plug}, which places ``Hyper Attention'' blocks at predetermined layers to run cross-attention. 

More recent models like Qwen3-VL~\cite{qwen3vl} similarly rely on a predetermined, hard-wired connection scheme. 
DEHVF~\cite{wei2025dynamic} further explores this direction by imposing a rigid, pre-defined ``one-to-one'' mapping between groups of vision and LLM layers. 
All these approaches lock the model into a manually configured, static ``straitjacket'' that predetermines information flow. 

The proposed CLI framework is designed to overcome key limitations of prior multimodal methods. Unlike DeepStack or CogVLM, our AGF module integrates information with selectivity and precision. 
Furthermore, though the concurrent work DEHVF employs a rigid group mapping, our framework empowers the LLM to learn this alignment by selectively querying the entire visual hierarchy. 
Similarly, in contrast to the hard-wired architectures of EVLM and mPLUG-Owl3, our framework provides on-demand access to the complete visual hierarchy at all injection points. This design enables the stable and dynamic many-to-many fusion essential for sophisticated multimodal reasoning.

\section{Method}

\begin{figure}[t]
\centering
\includegraphics[width=1.0\textwidth]{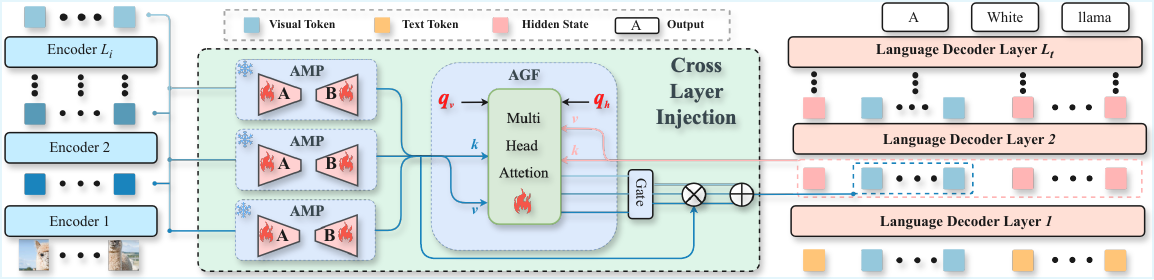}
\caption{An overview of the proposed Cross-Layer Injection (CLI) Framework. 
The left panel illustrates the overall ``many-to-many'' information flow, where hierarchical features from multiple vision encoder layers are dynamically injected into the LLM at multiple decoder layers. 
The right panel details the core \textbf{Adaptive Gating Fusion} (AGF) module that enables this process. 
Multi-Head Attention (MHA) first distills key information from both the incoming visual features and the current LLM hidden state. 
A gate controller then uses these distilled representations to compute a dynamic weight that governs the selective fusion of the new visual information. 
This gated process ensures the feature injection is both adaptive and context-sensitive, preventing the LLM from being overwhelmed by irrelevant data.}
\label{fig:architecture}
\end{figure}

This section details the Cross-Layer Injection (CLI) framework.
By replacing the conventional crude connections with a dynamic \textbf{many-to-many} bridge, CLI overcomes the information bottleneck that perceptually impoverishes VLMs. 
We first review the standard VLM pipeline in Sec.~\ref{Method:Preliminary} to contextualize its limitations.
Subsequently, Sec.~\ref{Method:Cross-Layer Injection} details the proposed mechanism. 
As shown in Fig.~\ref{fig:architecture}, it consists of two core components. 
The first component is \textbf{Adaptive Multi-Projection (AMP)}, a parameter-efficient module that uses LoRA to align multi-level visual features with the language space. 
The second, \textbf{Adaptive Gating Fusion (AGF)}, is a query-based attention gate that dynamically integrates these aligned features into the LLM during the decoding process.

\subsection{Preliminaries}
\label{Method:Preliminary}
The traditional VLM architecture (see Fig.~\ref{fig:traditional_mllm} (a)) comprises a visual encoder, a projector, and an LLM. 
First, the visual encoder extracts features from an input image $I$:
\begin{equation}
    V = \text{ImageEncoder}(I), \quad V \in \mathbb{R}^{N_i \times D_i},
\end{equation}
where $V$ consists of $N_i$ visual tokens, each with a dimension of $D_i$.
A projector module $P$, typically an MLP, then aligns these visual tokens with the language embedding space:
\begin{equation}
    \hat{V} = P(V) = \text{MLP}(V), \quad \hat{V} \in \mathbb{R}^{N_i \times D_t},
\end{equation}
where the output dimension $D_t$ matches the dimension of the text tokens.
Simultaneously, the input text, $T$, containing user instructions and prompts, is embedded into a sequence of $N_t$ text tokens.
The projected visual tokens $\hat{V}$ and the text tokens $\hat{T}$ are concatenated to form a unified multimodal context, $\mathbf{H} = [\hat{V}; \hat{T}]$, on which the LLM performs auto-regressive decoding: 
\begin{align}
    P(w_{t+1}|w_{1:t}, \mathbf{H}) &= \text{LLM}(w_{1:t}, \mathbf{H}).
\end{align}
This dominant paradigm's critical flaw, as argued in our introduction, is its reliance on only the final-layer visual features. This rigid one-to-one connection creates a severe information bottleneck by discarding the rich hierarchical information captured in earlier ViT layers.


\subsection{Many-to-Many Cross-Layer Injection}
\label{Method:Cross-Layer Injection}

To overcome this limitation, we introduce Many-to-Many Cross-Layer Injection, a framework that restores this symmetry by forging a dynamic many-to-many bridge between the vision encoder and the LLM decoder, as shown in Fig.~\ref{fig:architecture}. Instead of a single point of contact, CLI empowers the LLM to dynamically access and integrate multi-level visual features at various stages of its decoding process, conditioning its reasoning on a more comprehensive visual context.


Our method begins by extracting a hierarchical set of visual features. 
Instead of using only the final layer, we sample intermediate token matrices from $L_V$ distinct layers of the vision encoder, forming a collection $\mathbb{V}=\{V_l\}_{l=1}^{L_V}$. 
This strategy ensures that the model can access both low-level structural features from early layers and high-level semantic concepts from later ones. For instance, sampling from layers 1, 7, and 14 would produce the feature set $\mathbb{V}=\{V_1, V_7, V_{14}\}$.
To integrate these multi-level features during reasoning, we introduce cross-layer injection points at multiple layers within the LLM's decoder. 
This is operationalized through our novel \textbf{AMP} and \textbf{AGF}. 
At a specific layer in the LLM, these modules take the entire set of hierarchical visual features $\mathbb{V}$ as input, aggregate this information, and inject a synthesized representation. 
This allows the LLM to dynamically condition its output on the most relevant visual details, regardless of their layer of origin.

\vspace{1.0ex}
\noindent\textbf{Adaptive Multi Projection (AMP).} 
A central component of our mechanism is a specialized projection module designed to align multi-level visual features with the language embedding space. 
Aligning multi-level visual features presents a fundamental projection dilemma. 
On one hand, a single, rigid projector cannot reconcile the significant distributional variance across vision layers—from low-level textures to high-level semantics—leading to severe feature misalignment. 
On the other hand, the seemingly obvious alternative of training a dedicated projector for each layer is computationally prohibitive due to the extensive pretraining required.
Therefore, our solution is to make the projector \textit{adaptive} in a parameter-efficient manner. 
We employ LoRA to modify the behavior of the original, pre-trained projector. 
Specifically, for the visual tokens $V_k \in \mathbb{V}$ drawn from a sampled layer $k$, we augment the pre-trained MLP within the projector with a unique, layer-specific LoRA projector.
This adaptive projection is: 
\begin{align}
    \hat{V_k} = P(V_k) + \text{LoRA}(V_k) 
    = \text{MLP}(V_k) + B_kA_k(V_k), \quad V_k \in \mathbb{V},
\end{align}
where $A_k$ and $B_k$ are the low-rank matrices trained specifically for features from vision layer $k$.
This adaptive projection process is applied to each feature map $V_k \in \mathbb{V}$, yielding a new set of aligned token maps $\hat{\mathbb{V}} = \{\hat{V_k}\}_{k=1}^K$. 
Each component $\hat{V_k}$ within this set is now harmonized in the text embedding dimension, making the entire hierarchical collection ready for effective integration into the LLM.

\vspace{1.0ex}
\noindent\textbf{Adaptive Gating Fusion (AGF).}
With the visual tokens aligned, the next step is their effective integration. 
Rather than using a simple summation, which can disrupt the LLM's state, we propose an adaptive injection mechanism.
This approach acknowledges that the LLM's hidden state, $h$, already contains contextual information and thus requires a more nuanced update. 
Accordingly, for each injection layer $L_t$ in the LLM, a gating module is proposed to dynamically assess the relevance of the new visual features, $\hat{V}_k \in \hat{\mathbb{V}}$, based on the current decoding context $h_t$.
The gate's logic is driven by cross-attention. 
Two learnable query vectors, $q_v$ and $q_h$, act as probes to distill the essence of the new visual information and the existing hidden state:
\begin{align}
\hat{V}_{\text{att}} &= \text{MultiHeadAttention}(q_v, \hat{V}_k, \hat{V}_k),
\label{equ:att_v} \\
h_{\text{att}} &= \text{MultiHeadAttention}(q_h, h_t, h_t).
\label{equ:att_h}
\end{align}
The resulting context vectors are fused and passed through a gate controller—a linear layer followed by a Sigmoid activation—to yield a dynamic weight:
\begin{align}
    W &= \text{Sigmoid}(\text{Gate}([\hat{V}_{\text{att}}; h_{\text{att}}])),
\end{align}
where $W \in [0, 1]$.
This weight governs the selective update of the hidden state. 
To ensure precision, we use a binary ``mask'' that isolates the positions of visual tokens within $h_t$. 
The non-visual portions are preserved, while the visual portions are updated via a weighted sum with the new features:
\begin{equation}
    h'_t = h_t \odot (1-\text{mask}) + (h_t \odot \text{mask} + W \odot \hat{V}),
\end{equation}
where $\odot$ denotes element-wise product. 
This fusion process enriches the hidden state $h_t$ with a hierarchical representation of the visual input through processing all the $\hat{V}_k$ in $\hat{\mathbb{V}}$.
This gating and update cycle is repeated at designated injection points throughout the LLM's decoder. 
This method facilitates an iterative refinement of the model's visual understanding, allowing it to ``re-examine'' visual evidence at varying granularities throughout the generation process.

\section{Experiments}

This section presents a comprehensive empirical validation of our Cross-Layer Injection (CLI) framework. 
We first detail the experimental setup, including our implementation of CLI on two distinct VLM architectures and the benchmarks used for evaluation. 
Then, the main results are presented, demonstrating that CLI consistently and significantly outperforms strong baselines and competing fusion strategies across 28 diverse benchmarks. 
Finally, a series of in-depth ablation studies and analyses are conducted to dissect the specific contributions of CLI’s core components and validate our ``many-to-many'' design philosophy.


\subsection{Experiment Setup}
\noindent\textbf{Model Configuration.}
To demonstrate the versatility and general applicability of CLI, we integrate it into two distinct VLM architectures, LLaVA-OneVision \cite{DBLP:journals/tmlr/0080ZGZ00ZZL0L25} and LLaVA-1.5 \cite{DBLP:conf/cvpr/LiuLLL24}. The specific components for each are as follows:
\begin{itemize}
\item \textbf{CLI on LLaVA-OneVision}: This model uses a Qwen-2 series model~\cite{DBLP:journals/corr/abs-2407-10671} as the LLM backbone, a Siglip-so400m-patch14-384~\cite{DBLP:conf/iccv/ZhaiM0B23} vision encoder, and a two-layer MLP projector. 
To mitigate potential data contamination and ensure a fair evaluation, we initialize our fine-tuning from the public checkpoint released after its ``High-Quality Knowledge Learning'' stage.
\item \textbf{CLI on LLaVA-1.5}: 
This variant is built upon Vicuna-7B~\cite{vicuna2023} as the LLM and CLIP-Large~\cite{DBLP:conf/icml/RadfordKHRGASAM21} as the vision encoder, with a two-layer MLP projector using a GELU activation. 
Following the methodology of LLaVA-1.5, we fine-tune the model starting from the pre-trained checkpoint and using the identical training data.
\end{itemize}
We adopt the original training protocols from LLaVA-OneVision and LLaVA-1.5. 
Detailed hyper-parameters for our experiments are listed in Appendix~\ref{app:config}.

\vspace{1.0ex}
\noindent\textbf{Dataset Configuration.}
The foundation of our instruction tuning set consists of the single-image datasets from LLaVA-OneVision, which provide a broad base for general visual understanding (see Appendix~\ref{app:data} for details).

\vspace{1.0ex}
\noindent\textbf{Evaluation Benchmarks.}
To ensure a standardized and reproducible comparison, we evaluate the LLaVA-OneVision and LLaVA-1.5 models across a comprehensive suite of single-image benchmarks using the LMMs-Eval framework \cite{DBLP:conf/naacl/ZhangLZPCHLZYLL25}. 
These benchmarks are grouped into three primary categories to assess a wide range of capabilities (details can be found at Appendix~\ref{app:benchmark}):
\begin{itemize}
\item \textbf{Chart, Diagram, and Document Understanding}: AI2D \cite{DBLP:conf/eccv/KembhaviSKSHF16}, ChartQA \cite{DBLP:conf/acl/MasryLTJH22}, DocVQA \cite{DBLP:conf/wacv/MathewKJ21}, and InfoVQA \cite{DBLP:conf/wacv/MathewBTKVJ22}. 
These tasks demand fine-grained perception of text and structural elements.

\item \textbf{Perception and Multidisciplinary Reasoning}: MME \cite{DBLP:journals/corr/abs-2306-13394}, MMBench \cite{DBLP:conf/eccv/LiuDZLZZYWHLCL24}, MMVet \cite{DBLP:conf/icml/YuYLWL0WW24}, MathVerse \cite{DBLP:conf/eccv/ZhangJZLGQZLCQGL24}, MathVista \cite{DBLP:journals/corr/abs-2310-02255}, MMMU \cite{DBLP:conf/cvpr/YueNZ0LZSJRSWYY24}, GQA \cite{DBLP:conf/cvpr/HudsonM19}, OK-VQA \cite{DBLP:conf/cvpr/MarinoRFM19}, ScienceQA \cite{DBLP:conf/nips/LuMX0CZTCK22}, SEED-Bench \cite{DBLP:conf/acl/YingCWJWYSKYD25}, MM-Star \cite{DBLP:conf/nips/ChenLDZZCDWQLZ24}, and POPE \cite{DBLP:conf/emnlp/LiDZWZW23}. 
These benchmarks test abstract reasoning, knowledge integration, and resistance to hallucination.

\item \textbf{Real-world Understanding and Visual Chat}: RealworldQA \cite{Grok-1.5}, LLaVA-in-the-Wild \cite{DBLP:conf/cvpr/LiuLLL24}. 
These qualitative benchmarks evaluate the model's practical conversational and reasoning abilities in open-ended scenarios.
\end{itemize}

\noindent \textbf{Comparison Methods.}
We evaluate our method against two sets of competitors. 
First, to situate our model’s performance in the broader landscape, we compare it against SOTA VLMs at both similar  (IXC-2.5-7B~\cite{DBLP:journals/corr/abs-2407-03320}, InternVL-2-8B~\cite{DBLP:journals/corr/abs-2312-14238}) and larger scales (VILA-13B~\cite{DBLP:conf/cvpr/LinYP0SH24}, InternVL-2-26B~\cite{DBLP:journals/corr/abs-2312-14238}).
Additionally, to directly assess our fusion strategy, we conduct a controlled study comparing CLI against three fusion paradigms, all implemented on the same base models and trained on the identical dataset. 
Our primary comparison is against the \textbf{Baseline Projector}, the original single-layer projection method from the LLaVA architectures, which we re-trained to serve as a fair baseline (referred to as LLaVA-OV-0.5B/7B). 
We also compare against \textbf{DeepStack}~\cite{DBLP:conf/nips/MengYTDW0024}, a brute-force approach that partitions high-resolution visual tokens and injects these different token groups into different LLM layers, and \textbf{Shallow-Layer Injection (SLI)}, a rigid statically-wired method used by models like Qwen3-VL~\cite{qwen3vl} that creates a fixed one-to-one mapping from $n$ vision layers sampled at a uniform stride of 8 to the corresponding initial $n$ LLM decoder layers.


\subsection{Main Results}

We evaluate our CLI by incorporating it into LLaVA-OneVision and LLaVA-1.5 architectures. Quantitative results compared with SOTA methods and other fusion works are shown in Tabs.~\ref{Main_result_1} and~\ref{Main_result_2}.
Overall, these results show that our CLI framework significantly outperforms the Baseline Projector (labeled LLaVA-OV-0.5B and -7B) across both model scales. 
This validates our central hypothesis: empowering an LLM with dynamic, on-demand access to the full visual hierarchy unlocks a deeper level of multimodal understanding. 
Furthermore, the performance lift from CLI is substantially larger than that of other deep fusion strategies, and it is the only method to deliver robust gains over the baseline. 
DeepStack, with its context-blind addition of high-resolution patch tokens, often degrades performance, falling well below the baseline and demonstrating that unfiltered feature injection is disruptive.
Its strategy of partitioning visual tokens and injecting them into fixed layers via simple addition is non-selective; the LLM is forced to process pre-determined visual information regardless of its contextual relevance, which often disrupts its internal representations.
Shallow-Layer Injection delivers only marginal or inconsistent gains, failing to significantly improve upon the baseline because it confines multi-level visual data to the LLM's shallowest layers, leaving the deeper, reasoning-intensive layers perceptually impoverished. 
In stark contrast, CLI’s dynamic many-to-many architecture, governed by AGF, allows every designated LLM layer to query the entire visual hierarchy. 
This context-aware fusion enables a more sophisticated integration of visual evidence, leading to demonstrably superior performance.

\begin{table*}[t]
\caption{Performance evaluation of the proposed CLI framework across a suite of nine benchmarks. 
We integrate our proposed Cross-Layer Injection (CLI) framework into LLaVA-OV (0.5B, 7B) and compare against baselines and alternative fusion methods.
}
\vspace{-2ex} 
\label{Main_result_1}
\renewcommand\arraystretch{1.2}
\renewcommand\tabcolsep{4.0pt}
\centering
\resizebox{\linewidth}{!}{
\begin{tabular}{l|cccccccccc}
\toprule
\multicolumn{1}{c|}{\multirow{2}{*}{\textbf{Model}}} & AI2D & ChartQA & DocVQA & InfoVQA & RealWorldQA & LLaVA-W & POPE & OK-VQA & GQA & \multirow{2}{*}{\textbf{Sum}} \\ \cmidrule(lr){2-10}
\multicolumn{1}{c|}{} & test & test & val & val & test & test & test & val & test & \\ \midrule
VILA-13B \cite{DBLP:conf/cvpr/LinYP0SH24} & 57.6 & 32.8 & 20.0 & 10.0 & 41.9 & 58.6 & 0.4 & 1.6 & 17.6 & 240.5 \\
IXC-2.5-7B \cite{DBLP:journals/corr/abs-2407-03320} &39.1&81.2&90.3&68.1&57.5&63.2&88.5&29.2&57.7& 574.8 \\
InternVL-2-8B \cite{DBLP:journals/corr/abs-2312-14238} & 82.2 & 82.5 & 90.0 & 66.5 & 64.4 & 72.7 & 87.8 & 52.1 & 62.7 & 660.9 \\
InternVL-2-26B \cite{DBLP:journals/corr/abs-2312-14238} & 83.0 & 84.4 & 90.4 & 68.9 & 67.2 & 90.6 & 88.8 & 48.3 & 65.1 & 686.7 \\ \midrule \midrule
LLaVA-OV-0.5B & 56.5 & 64.5 & 64.1 & \textbf{47.5} & 56.0 & \textbf{61.7} & 88.8 & 48.4 & 53.7 & 541.2 \\
\rowcolor{mygray}\ \textit{w}/ DeepStack \cite{DBLP:conf/nips/MengYTDW0024} & 53.2\textsubscript{\color{mygreen}-3.3} & 54.0\textsubscript{\color{mygreen}-10.5} & 54.6\textsubscript{\color{mygreen}-9.5} & 41.0\textsubscript{\color{mygreen}-6.5} & 52.9\textsubscript{\color{mygreen}-3.1} & 57.1\textsubscript{\color{mygreen}-4.6} & 88.2\textsubscript{\color{mygreen}-0.6} & 46.2\textsubscript{\color{mygreen}-2.2} & 53.3\textsubscript{\color{mygreen}-0.4} & 490.5\textsubscript{\color{mygreen}-50.7} \\
\rowcolor{mygray}\ \textit{w}/ SLI \cite{qwen3vl} & \textbf{57.2}\textsubscript{\color{myred}+0.7} & 64.7\textsubscript{\color{myred}+0.2} & 63.6\textsubscript{\color{mygreen}-0.5} & 46.6\textsubscript{\color{mygreen}-0.9} & 55.1\textsubscript{\color{mygreen}-0.9} & 55.2\textsubscript{\color{mygreen}-6.5} & 89.0\textsubscript{\color{myred}+0.2} & \textbf{48.8}\textsubscript{\color{myred}+0.4} & 54.5\textsubscript{\color{myred}+0.8} & 534.7\textsubscript{\color{mygreen}-6.5} \\
\rowcolor{mygray}\ \textit{w}/ \textbf{CLI} & 56.7\textsubscript{\color{myred}+0.2} & \textbf{65.2}\textsubscript{\color{myred}+0.7} & \textbf{64.7}\textsubscript{\color{myred}+0.6} & 47.1\textsubscript{\color{mygreen}-0.4} & \textbf{56.7}\textsubscript{\color{myred}+0.7} & 61.1\textsubscript{\color{mygreen}-0.6} & \textbf{89.1}\textsubscript{\color{myred}+0.3} & 48.6\textsubscript{\color{myred}+0.2} & \textbf{53.7} & 542.9\textsubscript{\color{myred}+1.7} \\
LLaVA-OV-7B & 77.5 & 78.5 & 82.5 & 69.5 & \textbf{68.4} & 68.0 & 88.6 & 58.5 & 59.4 & 650.9 \\
\rowcolor{mygray}\ \textit{w}/ DeepStack \cite{DBLP:conf/nips/MengYTDW0024} & 76.0\textsubscript{\color{mygreen}-1.5} & 63.8\textsubscript{\color{mygreen}-14.7} & 72.6\textsubscript{\color{mygreen}-9.9} & 62.2\textsubscript{\color{mygreen}-7.3} & 61.4\textsubscript{\color{mygreen}-7.0} & 70.3\textsubscript{\color{myred}+2.3} & 88.7\textsubscript{\color{myred}+0.1} & 58.3\textsubscript{\color{mygreen}-0.2} & 58.5\textsubscript{\color{mygreen}-0.9} & 611.8\textsubscript{\color{mygreen}-39.1} \\
\rowcolor{mygray}\ \textit{w}/ SLI \cite{qwen3vl} & 77.6\textsubscript{\color{myred}+0.1} & 77.3\textsubscript{\color{mygreen}-1.2} & 79.9\textsubscript{\color{mygreen}-2.6} & 67.6\textsubscript{\color{mygreen}-1.9} & 68.2\textsubscript{\color{mygreen}-0.2} & 68.5\textsubscript{\color{myred}+0.5} & \textbf{89.0}\textsubscript{\color{myred}+0.4} & 58.4\textsubscript{\color{mygreen}-0.1} & 59.3\textsubscript{\color{mygreen}-0.1} & 645.8\textsubscript{\color{mygreen}-5.1} \\
\rowcolor{mygray}\ \textit{w}/ \textbf{CLI} & \textbf{77.9}\textsubscript{\color{myred}+0.4} & \textbf{78.7}\textsubscript{\color{myred}+0.2} & \textbf{82.8}\textsubscript{\color{myred}+0.3} & \textbf{70.5}\textsubscript{\color{myred}+1.0} & \textbf{68.4} & \textbf{74.5}\textsubscript{\color{myred}+6.5} & 88.9\textsubscript{\color{myred}+0.3} & \textbf{59.2}\textsubscript{\color{myred}+0.7} & \textbf{59.7}\textsubscript{\color{myred}+0.3} & \textbf{660.6}\textsubscript{\color{myred}+9.7} \\ 
\bottomrule
\end{tabular}
}
\end{table*}

\begin{figure}[t]
\vspace{-2ex}
\includegraphics[width=1.0\linewidth]{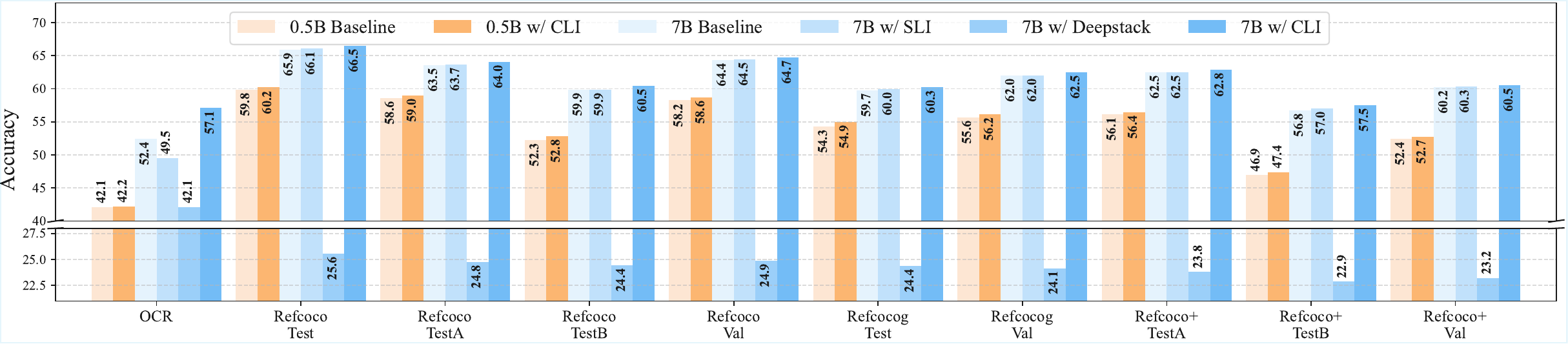}
\vspace{-5ex}
\caption{Performance gains from CLI on fine-grained visual reasoning. 
On both OCR and a comprehensive suite of visual grounding benchmarks (RefCOCO/+/g), CLI delivers consistent improvements for both the 0.5B and 7B models. 
}
\label{fig:grounding_ocr}
\end{figure}

Specifically, the benefits of CLI's architecture are evident across diverse task categories. 
For \textbf{fine-grained document understanding} on benchmarks like AI2D, ChartQA, DocVQA, and InfoVQA, CLI delivers cumulative gains of +1.1\% and +1.9\% over the Baseline Projector for the 0.5B and 7B models, respectively, by granting access to early-layer ViT features rich in essential textural and structural details. 
This strength extends to \textbf{complex multimodal reasoning}, where the impressive +6.5\% gain on LLaVA-in-the-Wild with open-ended questions for the 7B model showcases how the AGF mechanism allows the LLM to dynamically synthesize information from the entire visual hierarchy—a capability static methods lack. 
Consistent gains on other reasoning-heavy benchmarks like MME (+3.3\% on 7B) and SEED-Bench (+1.1\% on 0.5B) further confirm that rich, multi-level visual input is a critical component for high-level cognitive tasks.
This enhancement further allows our CLI-enhanced models to compete with much larger systems. 
Notably, our LLaVA-OV-7B with CLI achieves a total score of 660.6, closing the performance gap to reach near-parity with the larger InternVL-2-8B model (660.9).

\begin{table*}[t]
\caption{Main results on benchmarks for perception, multidisciplinary reasoning, and integrated capabilities. 
We compare our CLI-enhanced models against their baselines and alternative fusion methods on nine challenging benchmarks. 
}
\vspace{-2ex}
\label{Main_result_2}
\renewcommand\arraystretch{1.2}
\renewcommand\tabcolsep{4.0pt}
\centering
\resizebox{\linewidth}{!}{
\begin{tabular}{l|cccccccccc}
\toprule
\multicolumn{1}{c|}{\multirow{2}{*}{\textbf{Model}}} 
 & MathVerse & MathVista & MMBench & MME & MMStar & MMMU & MMVet & SeedBench & ScienceQA & \multirow{2}{*}{\textbf{Sum}} \\ \cmidrule(lr){2-10} 
 & mini-vision & testmini & en-dev & test & test & val & test & image & test &  \\ \midrule
VILA-13B \cite{DBLP:conf/cvpr/LinYP0SH24} & 26.4 & 33.7 & 57.9 & 45.2 & 32.8 & 31.5 & 31.2 & 66.5 & 71.0 & 396.2 \\
IXC-2.5-7B \cite{DBLP:journals/corr/abs-2407-03320} & 13.03 & 60.4 & 66.1 & 90.3 & 14.8 & 33.1 & 46.3 & 47.1 & 22.0 & 380.1 \\
InternVL-2-8B \cite{DBLP:journals/corr/abs-2312-14238} & 31.2 & 63.0 & 81.7 & 93.0 & 58.7 & 48.5 & 53.6 & 76.1 & 97.0 & 602.8 \\
InternVL-2-26B \cite{DBLP:journals/corr/abs-2312-14238} & 27.3 & 61.1 & 81.6 & 94.9 & 60.4 & 47.5 & 53.1 & 76.7 & 97.5 & 600.1 \\ \midrule \midrule
LLaVA-OV-0.5B & 17.7 & 36.3 & 45.8 & 63.2 & 38.5 & 30.7 & 28.0 & 65.6 & 65.2 & 391.0 \\
\rowcolor{mygray}\ \textit{w}/ DeepStack \cite{DBLP:conf/nips/MengYTDW0024} & 17.8\textsubscript{\color{myred}+0.1} & 32.5\textsubscript{\color{mygreen}-3.8} & \textbf{51.2}\textsubscript{\color{myred}+5.4} & 59.8\textsubscript{\color{mygreen}-3.4} & 38.5 & \textbf{31.6}\textsubscript{\color{myred}+0.9} & 23.1\textsubscript{\color{mygreen}-4.9} & 62.9\textsubscript{\color{mygreen}-2.7} & 63.5\textsubscript{\color{mygreen}-1.7} & 380.9\textsubscript{\color{mygreen}-10.1} \\
\rowcolor{mygray}\ \textit{w}/ SLI \cite{qwen3vl} & 18.0\textsubscript{\color{myred}+0.3} & \textbf{37.9}\textsubscript{\color{myred}+1.6} & 47.3\textsubscript{\color{myred}+1.5} & 61.1\textsubscript{\color{mygreen}-2.1} & 42.6\textsubscript{\color{myred}+4.1} & 30.2\textsubscript{\color{mygreen}-0.5} & 26.1\textsubscript{\color{mygreen}-1.9} & 65.9\textsubscript{\color{myred}+0.3} & 64.6\textsubscript{\color{mygreen}-0.6} & 393.7\textsubscript{\color{myred}+2.7} \\
\rowcolor{mygray}\ \textit{w}/ \textbf{CLI} & \textbf{18.3}\textsubscript{\color{myred}+0.6} & 37.1\textsubscript{\color{myred}+0.8} & 48.3\textsubscript{\color{myred}+2.5} & \textbf{63.3}\textsubscript{\color{myred}+0.1} & \textbf{39.3}\textsubscript{\color{myred}+0.8} & 30.1\textsubscript{\color{mygreen}-0.6} & 27.0\textsubscript{\color{mygreen}-1.0} & \textbf{66.7}\textsubscript{\color{myred}+1.1} & \textbf{65.5}\textsubscript{\color{myred}+0.3} & \textbf{395.6}\textsubscript{\color{myred}+4.6} \\
LLaVA-OV-7B & \textbf{28.8} & 55.4 & 79.2 & 85.1 & 54.1 & 47.0 & \textbf{44.1} & 76.1 & \textbf{86.4} & 556.2 \\
\rowcolor{mygray}\ \textit{w}/ DeepStack \cite{DBLP:conf/nips/MengYTDW0024} & 27.0\textsubscript{\color{mygreen}-1.8} & 55.5\textsubscript{\color{myred}+0.1} & 78.1\textsubscript{\color{mygreen}-1.1} & 84.1\textsubscript{\color{mygreen}-1.0} & 50.8\textsubscript{\color{mygreen}-3.3} & 45.6\textsubscript{\color{mygreen}-1.4} & 40.3\textsubscript{\color{mygreen}-3.8} & 74.3\textsubscript{\color{mygreen}-1.8} & 83.6\textsubscript{\color{mygreen}-2.8} & 539.3\textsubscript{\color{mygreen}-16.9} \\
\rowcolor{mygray}\ \textit{w}/ SLI \cite{qwen3vl} & 25.9\textsubscript{\color{mygreen}-2.9} & \textbf{57.8}\textsubscript{\color{myred}+2.4} & \textbf{79.6}\textsubscript{\color{myred}+0.4} & 84.4\textsubscript{\color{mygreen}-0.7} & 55.8\textsubscript{\color{myred}+1.7} & 44.7\textsubscript{\color{mygreen}-2.3} & 40.6\textsubscript{\color{mygreen}-3.5} & \textbf{76.5}\textsubscript{\color{myred}+0.4} & 85.8\textsubscript{\color{mygreen}-0.6} & 551.1\textsubscript{\color{mygreen}-5.1} \\
\rowcolor{mygray}\ \textit{w}/ \textbf{CLI} & 27.1\textsubscript{\color{mygreen}-1.7} & 55.8\textsubscript{\color{myred}+0.4} & 78.9\textsubscript{\color{mygreen}-0.3} & \textbf{88.4}\textsubscript{\color{myred}+3.3} & \textbf{56.5}\textsubscript{\color{myred}+2.4} & \textbf{47.6}\textsubscript{\color{myred}+0.6} & 43.8\textsubscript{\color{mygreen}-0.3} & 75.7\textsubscript{\color{mygreen}-0.4} & 85.6\textsubscript{\color{mygreen}-0.8} & 559.4\textsubscript{\color{myred}+3.2} \\ 
\bottomrule
\end{tabular}
}
\vspace{-2ex}
\end{table*}

To further challenge the framework on tasks that depend critically on multi-level visual information, we evaluated it on a separate suite of fine-grained benchmarks: Optical Character Recognition (OCR) and visual grounding (as shown in Fig.~\ref{fig:grounding_ocr}). 
These tasks act as a stress-test for vision-language fusion, as they require the model to resolve fine character strokes (for OCR) or synthesize high-level semantic context with precise low-level localization (for visual grounding). 
Here, CLI's superiority is unambiguous. Its ability to tap into early-layer ViT features for high-fidelity detail proves decisive, yielding a remarkable \textbf{4.7\%} improvement for the 7B model on OCR. 
Similarly, it excels at grounding by allowing the LLM to dynamically draw on both deep ViT layers for context and shallow layers for spatial detail, leading to robust gains across \textbf{all nine} grounding benchmarks. 
This stands in stark contrast to the inconsistent results of competing methods as shown in Fig.~\ref{fig:grounding_ocr}, confirming that a dynamic, many-to-many architecture can reliably satisfy these complex visual demands.

\begin{wrapfigure}{t}{0.35\linewidth}
\centering  
\vspace{-4ex}
\includegraphics[width=1.0\linewidth]{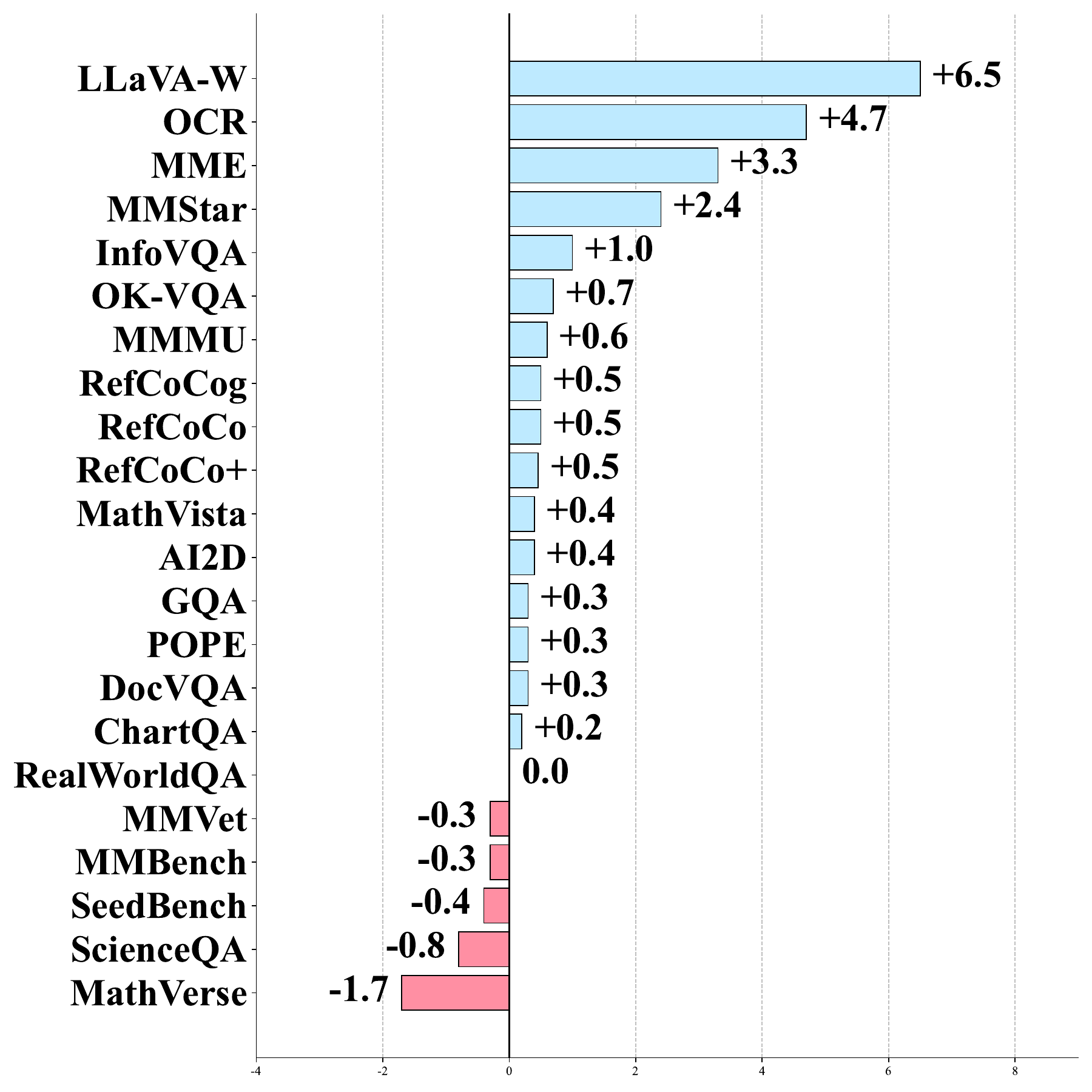}
\vspace{-5ex}
\caption{Performance Gains of CLI on LLaVA-OV-7B. 
Absolute performance delta compared to the baseline across 28 benchmarks.
}
\label{fig:win_dataset}
\vspace{-5ex}
\end{wrapfigure}
To offer a holistic and intuitive validation of our framework, Fig.~\ref{fig:win_dataset} displays CLI's performance gains over the baseline across 28 benchmarks, sorted by impact. CLI achieves a \textbf{+19.06} aggregate score increase, with the largest gains in challenging domains like open-ended reasoning (LLaVA-W: +6.5) and fine-grained perception (OCR: +4.7). On the few tasks where low-level features could be detrimental, CLI's gating mechanism effectively mitigates interference, confirming its architectural robustness.

In addition, to validate its generalizability, we ported CLI to the LLaVA-1.5 architecture, which uses a different LLM, vision encoder, and pre-training pipeline.
As reported in Tab.~\ref{tab:Main_result_llava1.5}, CLI delivers comprehensive performance improvements to this architecture as well.
The gains on LLaVA-1.5 are even more pronounced than on LLaVA-OV (+7.5 and +8.6 partial sum improvements vs. +9.7 and +3.2). This suggests a deeper advantage of our framework: CLI can act as a compensatory mechanism, mitigating pre-existing weaknesses in a base model's original visual-language alignment.
By forging a richer, more dynamic connection, CLI establishes a more robust multimodal foundation, highlighting its broad utility as a plug-and-play enhancement.
(A comparison with DeepStack and SLI can be found in Appendix~\ref{app:llava15_deepstack}.)

\begin{table*}[t]
\caption{
To validate the architecture-agnostic nature of our method, we integrated CLI into the LLaVA-1.5 architecture. 
The CLI-enhanced model demonstrates consistent and broad performance gains across a wide range of tasks, confirming the robustness and general applicability of our framework.
}
\vspace{-2ex}
\label{tab:Main_result_llava1.5}
\renewcommand\arraystretch{1.2}
\renewcommand\tabcolsep{4.0pt}
\centering
\resizebox{\linewidth}{!}{
\begin{tabular}{c|ccccccccc|c}
\toprule
\textbf{Model} & AI2D & ChartQA & DocVQA & InfoVQA & RealWorldQA & LLaVA-W & POPE & OK-VQA & GQA & \textbf{Partial Sum} \\
\midrule
LLaVA-1.5-7B & 66.3 & 38.9 & 32.2 & 26.8 & 54.5 & 62.9 & 87.0 & 41.8 & 57.6 & 468.0 \\
\rowcolor{mygray}
\ \textit{w}/ \textbf{CLI} & 65.7\textsubscript{\color{mygreen}-0.6} & 39.6\textsubscript{\color{myred}+0.7} & 32.4\textsubscript{\color{myred}+0.2} & 26.3\textsubscript{\color{mygreen}-0.5} & 54.6\textsubscript{\color{myred}+0.1} & 65.9\textsubscript{\color{myred}+3.0} & 86.4\textsubscript{\color{mygreen}-0.6} & 47.0\textsubscript{\color{myred}+5.2} & 57.6 & 475.5\textsubscript{\color{myred}+7.5} \\
\midrule

\textbf{Model} & MathVerse & MathVista & MMBench & MME & MMStar & MMMU & MMVet & SeedBench & ScienceQA & \textbf{Partial Sum} \\
\midrule
LLaVA-1.5-7B & 17.5 & 34.4 & 64.3 & 79.9 & 37.2 & 34.5 & 31.2 & 61.9 & 72.9 & 433.8 \\
\rowcolor{mygray}
\ \textit{w}/ \textbf{CLI} & 18.3\textsubscript{\color{myred}+0.8} & 35.6\textsubscript{\color{myred}+1.2} & 66.3\textsubscript{\color{myred}+2.0} & 79.4\textsubscript{\color{mygreen}-0.5} & 38.5\textsubscript{\color{myred}+1.3} & 35.7\textsubscript{\color{myred}+1.2} & 34.5\textsubscript{\color{myred}+3.3} & 61.9 & 72.2\textsubscript{\color{mygreen}-0.7} & 442.4\textsubscript{\color{myred}+8.6} \\
\bottomrule
\end{tabular}
}
\vspace{-2ex}
\end{table*}

\begin{table*}[t]
\vspace{-1ex}
\caption{Ablation study of CLI components on LLaVA-OV-0.5B. 
Incrementally adding the AMP and AGF modules demonstrates the critical role of the gating mechanism and reveals a strong synergy between both components.}
\vspace{-2ex}
\label{ablation_result_1}
\renewcommand\arraystretch{1.2}
\renewcommand\tabcolsep{4.0pt}
\centering
\resizebox{\linewidth}{!}{
\begin{tabular}{l|ccccccccc|cc}
\toprule
\multicolumn{1}{c|}{\multirow{2}{*}{\textbf{Variant}}} & AI2D & ChartQA & DocVQA   & InfoVQA  & MathVerse   & MathVista & MMBench & MME  & MMMU & \multirow{2}{*}{\textbf{Sum}} & \multirow{2}{*}{\textbf{Para}} \\ 
\cmidrule(lr){2-10}
        & test & test    & val/test & val/test & mini-vision & testmini  & en-dev  & test & val & &  \\ \midrule 
Baseline   &48.87&55.88&58.87&42.21&16.71&29.40&28.52&56.38&29.88 & 366.72 & 100.00\% \\ 
\textit{w/} AMP   &49.13&55.76&59.19&42.75&17.04&29.60&28.26&56.28&29.88 & 367.89 & 102.25\% \\
\textit{w/} AGF       &48.67&56.60&59.47&42.48&17.01&29.10&30.58&56.47&30.22  & 370.60 & 102.12\% \\
\textit{w/} AMP + AGF       & 48.64&55.40&59.32&42.52&17.85&30.00&30.41&56.71&30.66  & 371.51 & 104.37\% \\
\midrule
\textit{w/} Full AMP   &49.38&56.12&59.27&42.38&17.13&28.50&29.12&56.07&29.77 & 367.74 & 107.95\% \\
\textit{w/} Full AMP + AGF       &\textbf{51.30}&\textbf{62.72}&\textbf{60.76}&\textbf{44.71}&\textbf{18.06}&\textbf{33.30}&\textbf{45.79}&\textbf{58.49}&\textbf{31.88} & \textbf{407.01} & 110.07\% \\ 
\bottomrule
\end{tabular}
}
\vspace{-2ex}
\end{table*}

\subsection{Ablation Studies}

We conducted ablation studies on LLaVA-OneVision-0.5B to analyze CLI's components and injection strategy. 
For efficiency, these experiments used a 50\% subset of the instruction tuning data, a configuration justified by our scalability analysis (Appendix~\ref{app:data_size}). 
Key findings on component roles, injection density, and the AGF's internal query mechanism are summarized below, with full details provided in Appendix~\ref{app:ablation}.

\vspace{1.0ex}
\noindent\textbf{Dissecting the Contributions of CLI Components.}
A component-wise ablation (Tab.~\ref{ablation_result_1}) reveals the critical, synergistic roles of AMP and AGF. 
Injecting multi-level features using only AMP provides marginal gains (367.89 vs. 366.72 baseline), confirming that simply flooding the LLM with unfiltered information is ineffective. 
In stark contrast, incorporating the AGF gating module alone yields a significant performance uplift (370.60), decisively validating our core hypothesis: the ability to dynamically select visual information is the most critical factor for successful fusion.
The fine-tuned CLI framework, combining both modules, achieves further gains (371.51), demonstrating that AMP harmonizes the hierarchical features, providing a cleaner input for AGF to effectively gate. 
While a fully fine-tuned projector paired with AGF yields the highest absolute score (407.01), our LoRA-based AMP provides a much more compelling balance of performance and parameter efficiency (104.37\% vs. 110.07\% total params), justifying its use in our framework.

\begin{wrapfigure}{r}{0.4\linewidth}
\centering  
\vspace{-5ex}
\includegraphics[width=1.0\linewidth]{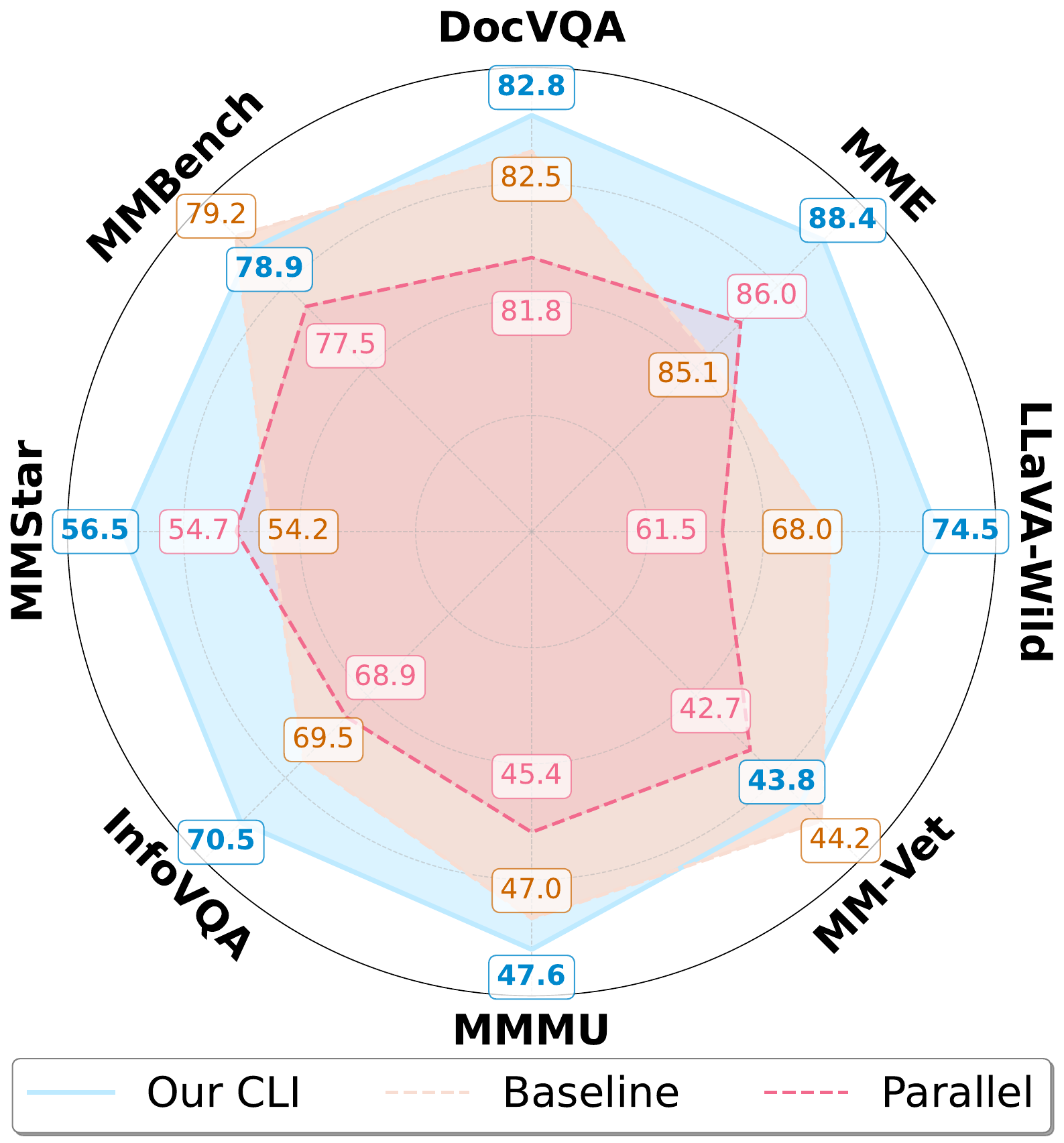}
\caption{
\textbf{Architectural Effectiveness of CLI vs. Static Parallel Fusion.}
}
\label{fig:7b_parallel}
\vspace{-4ex}
\end{wrapfigure}

\vspace{1.0ex}
\noindent\textbf{Impact of Injection Density.}
We next ablate the injection strategy to validate our many-to-many design (see Fig.~\ref{fig:layer_selection} in Appendix ~\ref{app:ablation}). 
Comparing various injection densities, we find that a single-point injection consistently underperforms, reaffirming that it creates an insurmountable information bottleneck.
Conversely, a high-density strategy—injecting features frequently at multiple layers—achieves the best overall performance, validating our principle that on-demand access to the full visual hierarchy is essential. 
Intriguingly, a medium-density configuration underperforms a sparser one, suggesting a trade-off where intermittent updates may introduce disruptive cognitive overhead without the benefit of the near-continuous context provided by a high-density approach.

\noindent\textbf{Impact of Architectural Effectiveness.}
To disentangle architectural benefits from parameter count, we conducted a controlled experiment with an equal-parameter setup. We designed a `Parallel' model with a parameter count comparable to CLI but implemented a static, one-to-one injection scheme ($i$-th vision to $i$-th decoder layer). 
As shown in Fig.~\ref{fig:7b_parallel}, this Parallel model frequently underperformed the baseline, dropping 6.5 points on LLaVA-Wild and yielding no significant gains on MME or MMStar. In stark contrast, CLI's dynamic architecture improved the LLaVA-Wild score by 6.5 points, creating a performance gap of over 13 points against the Parallel model, while also achieving consistent gains on other benchmarks. This result strongly indicates that CLI's performance gains stem from its architectural effectiveness, not merely an increased parameter count, thereby validating the design's complexity.

\begin{figure}[t]
\includegraphics[width=1.0\linewidth]{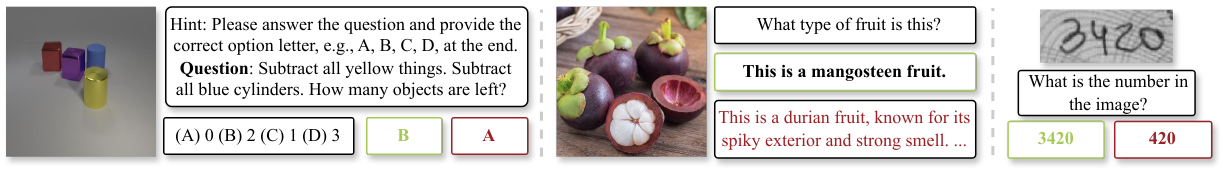}
\vspace{-5ex}
\caption{
Qualitative comparison of CLI on the LLaVA-OV-7B model across diverse benchmarks. 
Outputs from Baseline Projector are shown in \textcolor[HTML]{9B2226}{red}, while outputs from CLI are in \textcolor[HTML]{A7C957}{green}. 
The examples demonstrate that by integrating multi-level visual information, our model achieves more accurate perception and reasoning across various tasks, including MM-Star, LLaVA-in-the-Wild, and OCR-Bench.}
\vspace{-2ex}
\label{fig:visualization}
\end{figure}

\subsection{Qualitative and Efficiency Analysis}

\noindent\textbf{Qualitative Analysis.}
To offer insight into both the practical benefits and internal dynamics of our framework, we present qualitative and mechanistic visualizations. 
The qualitative examples in Fig.~\ref{fig:visualization} highlight CLI's superior performance across diverse tasks. 
In complex compositional reasoning (MM-Star), fine-grained recognition (LLaVA-in-the-Wild), and challenging OCR, our model demonstrates more nuanced and accurate perception than the baseline. 
For instance, it correctly identifies a mangosteen where the baseline sees a durian, and reads the entire number `\textbf{3420}' where the baseline outputs a fragment, showcasing an ability to integrate both fine-grained details and global context.
Additional visualizations and failure case analyses are provided in Appendix~\ref{app_qualitative}.

The mechanism enabling this is revealed by visualizing the learned gating weights (Fig.~\ref{fig:attention_sparsity} (b)). The heatmap uncovers a complex, non-parallel “criss-crossed” information flow, providing compelling evidence for our many-to-many fusion thesis. 
It demonstrates an evolving demand for visual information throughout the LLM's decoding process: shallow decoder layers query early vision layers to ground basic concepts, while deeper layers tasked with complex reasoning learn to query the entire visual hierarchy.
This allows them to simultaneously “zoom in” on fine details (crucial for OCR) and “zoom out” for global context (vital for reasoning).
Together, these visualizations confirm that granting the LLM dynamic, on-demand access to the full visual hierarchy is essential for sophisticated multimodal understanding and validates the efficacy of CLI's design.

\noindent\textbf{Computational Efficiency.}
To quantify the resource implications of our framework, we report the computational costs for both training and inference in Tab.~\ref{tab:costs}. 
The results indicate that CLI introduces a modest and manageable computational overhead compared to the baseline. The memory overhead, in particular, is remarkably marginal, with the inference footprint increasing by a negligible \underline{1.3\%}. This minimal impact is a critical advantage, demonstrating that our framework can be deployed without requiring significantly more VRAM. 
\begin{wraptable}{r}{0.55\columnwidth}
\vspace{-4ex}
\caption{Computational Costs per sample in MMBench.
(Memory indicates the Memory increases during forward).}
\resizebox{1.0\linewidth}{!}{
\begin{tabular}{lccccc}
\hline
\multirow{2}{*}{Method} & \multicolumn{3}{c}{Inference} & \multicolumn{2}{c}{Train} \\
 & Time & Memory & Flops & Memory & Flops \\ \hline
Baseline & 0.66 s & 241.2 MB & 3.7T & 251.3 MB & 5.46T \\
CLI & 0.77 s & 244.4 MB & 5.0T & 277.6 MB & 7.42T \\ \hline
\end{tabular}}
\label{tab:costs}
\end{wraptable}
This slight increase in computational demand is decisively justified by the substantial performance improvements it unlocks, including the \underline{+19.06} aggregate score gain shown in Fig.~\ref{fig:win_dataset}. 
This analysis thus confirms a highly favorable trade-off and substantiates our claim that CLI is a parameter-efficient method with minimal memory impact.

\section{Conclusion}
This paper introduces CLI, a framework that resolves the information bottleneck in VLMs by creating a dynamic ``many-to-many'' bridge between vision and language hierarchies. 
By granting the LLM on-demand access to all visual layers, CLI achieves significant performance gains across 28 diverse benchmarks. Our work validates that dynamic, selective fusion is critical for sophisticated reasoning, establishing CLI as a scalable paradigm for deeper multimodal understanding.

\section{Acknowledgements}
This study is supported by grants from the New Generation Artificial Intelligence-National Science and Technology Major Project (No.2025ZD0123005), National Natural Science Foundation of China (No. U22A2097, No.62425208, No. 62402094), and Fundamental Research Funds for the Central Universities.
In addition, this study is supported by Ant Group Research Intern Program

\bibliographystyle{splncs04}
\bibliography{eccv}

@String(CVPR  = {IEEE Conf. Comput. Vis. Pattern Recog.})

@String(ICCV  = {Int. Conf. Comput. Vis.})

@String(ECCV  = {Eur. Conf. Comput. Vis.})

@String(NeurIPS = {Adv. Neural Inform. Process. Syst.})

@String(ICML  = {Int. Conf. Mach. Learn.})

@String(ICLR  = {Int. Conf. Learn. Represent.})

@String(CVPR  = {CVPR})

@String(ICCV  = {ICCV})

@String(ECCV  = {ECCV})

@String(NeurIPS = {NeurIPS})

@String(ICML  = {ICML})

@String(ICLR  = {ICLR})

@inproceedings{DBLP:conf/icml/0008LSH23,
  author       = {Junnan Li and
                  Dongxu Li and
                  Silvio Savarese and
                  Steven C. H. Hoi},
  editor       = {Andreas Krause and
                  Emma Brunskill and
                  Kyunghyun Cho and
                  Barbara Engelhardt and
                  Sivan Sabato and
                  Jonathan Scarlett},
  title        = {{BLIP-2:} Bootstrapping Language-Image Pre-training with Frozen Image
                  Encoders and Large Language Models},
  booktitle    = {International Conference on Machine Learning, {ICML} 2023, 23-29 July
                  2023, Honolulu, Hawaii, {USA}},
  series       = {Proceedings of Machine Learning Research},
  volume       = {202},
  pages        = {19730--19742},
  publisher    = {{PMLR}},
  year         = {2023},
}

@article{DBLP:journals/tmlr/0080ZGZ00ZZL0L25,
  author       = {Bo Li and
                  Yuanhan Zhang and
                  Dong Guo and
                  Renrui Zhang and
                  Feng Li and
                  Hao Zhang and
                  Kaichen Zhang and
                  Peiyuan Zhang and
                  Yanwei Li and
                  Ziwei Liu and
                  Chunyuan Li},
  title        = {LLaVA-OneVision: Easy Visual Task Transfer},
  journal      = {Trans. Mach. Learn. Res.},
  volume       = {2025},
  year         = {2025},
}

@article{DBLP:journals/corr/abs-2407-10671,
  author       = {An Yang and
                  Baosong Yang and
                  Binyuan Hui and
                  Bo Zheng and
                  Bowen Yu and
                  Chang Zhou and
                  Chengpeng Li and
                  Chengyuan Li and
                  Dayiheng Liu and
                  Fei Huang and
                  Guanting Dong and
                  Haoran Wei and
                  Huan Lin and
                  Jialong Tang and
                  Jialin Wang and
                  Jian Yang and
                  Jianhong Tu and
                  Jianwei Zhang and
                  Jianxin Ma and
                  Jianxin Yang and
                  Jin Xu and
                  Jingren Zhou and
                  Jinze Bai and
                  Jinzheng He and
                  Junyang Lin and
                  Kai Dang and
                  Keming Lu and
                  Keqin Chen and
                  Kexin Yang and
                  Mei Li and
                  Mingfeng Xue and
                  Na Ni and
                  Pei Zhang and
                  Peng Wang and
                  Ru Peng and
                  Rui Men and
                  Ruize Gao and
                  Runji Lin and
                  Shijie Wang and
                  Shuai Bai and
                  Sinan Tan and
                  Tianhang Zhu and
                  Tianhao Li and
                  Tianyu Liu and
                  Wenbin Ge and
                  Xiaodong Deng and
                  Xiaohuan Zhou and
                  Xingzhang Ren and
                  Xinyu Zhang and
                  Xipin Wei and
                  Xuancheng Ren and
                  Xuejing Liu and
                  Yang Fan and
                  Yang Yao and
                  Yichang Zhang and
                  Yu Wan and
                  Yunfei Chu and
                  Yuqiong Liu and
                  Zeyu Cui and
                  Zhenru Zhang and
                  Zhifang Guo and
                  Zhihao Fan},
  title        = {Qwen2 Technical Report},
  journal      = {CoRR},
  volume       = {abs/2407.10671},
  year         = {2024},
}

@inproceedings{DBLP:conf/iccv/ZhaiM0B23,
  author       = {Xiaohua Zhai and
                  Basil Mustafa and
                  Alexander Kolesnikov and
                  Lucas Beyer},
  title        = {Sigmoid Loss for Language Image Pre-Training},
  booktitle    = {{IEEE/CVF} International Conference on Computer Vision, {ICCV} 2023,
                  Paris, France, October 1-6, 2023},
  pages        = {11941--11952},
  publisher    = {{IEEE}},
  year         = {2023},
}

@inproceedings{DBLP:conf/cvpr/LiuLLL24,
  author       = {Haotian Liu and
                  Chunyuan Li and
                  Yuheng Li and
                  Yong Jae Lee},
  title        = {Improved Baselines with Visual Instruction Tuning},
  booktitle    = {{IEEE/CVF} Conference on Computer Vision and Pattern Recognition,
                  {CVPR} 2024, Seattle, WA, USA, June 16-22, 2024},
  pages        = {26286--26296},
  publisher    = {{IEEE}},
  year         = {2024},
}

@inproceedings{DBLP:conf/icml/RadfordKHRGASAM21,
  author       = {Alec Radford and
                  Jong Wook Kim and
                  Chris Hallacy and
                  Aditya Ramesh and
                  Gabriel Goh and
                  Sandhini Agarwal and
                  Girish Sastry and
                  Amanda Askell and
                  Pamela Mishkin and
                  Jack Clark and
                  Gretchen Krueger and
                  Ilya Sutskever},
  title        = {Learning Transferable Visual Models From Natural Language Supervision},
  booktitle    = {Proceedings of the 38th International Conference on Machine Learning,
                  {ICML} 2021, 18-24 July 2021, Virtual Event},
  series       = {Proceedings of Machine Learning Research},
  volume       = {139},
  pages        = {8748--8763},
  publisher    = {{PMLR}},
  year         = {2021},
}

@misc{vicuna2023,
    title = {Vicuna: An Open-Source Chatbot Impressing GPT-4 with 90\%* ChatGPT Quality},
    url = {https://lmsys.org/blog/2023-03-30-vicuna/},
    author = {Chiang, Wei-Lin and Li, Zhuohan and Lin, Zi and Sheng, Ying and Wu, Zhanghao and Zhang, Hao and Zheng, Lianmin and Zhuang, Siyuan and Zhuang, Yonghao and Gonzalez, Joseph E. and Stoica, Ion and Xing, Eric P.},
    month = {March},
    year = {2023}
}

@inproceedings{DBLP:conf/eccv/KembhaviSKSHF16,
  author       = {Aniruddha Kembhavi and
                  Mike Salvato and
                  Eric Kolve and
                  Min Joon Seo and
                  Hannaneh Hajishirzi and
                  Ali Farhadi},
  title        = {A Diagram is Worth a Dozen Images},
  booktitle    = {Computer Vision - {ECCV} 2016 - 14th European Conference, Amsterdam,
                  The Netherlands, October 11-14, 2016, Proceedings, Part {IV}},
  series       = {Lecture Notes in Computer Science},
  volume       = {9908},
  pages        = {235--251},
  publisher    = {Springer},
  year         = {2016},
}

@inproceedings{DBLP:conf/acl/MasryLTJH22,
  author       = {Ahmed Masry and
                  Do Xuan Long and
                  Jia Qing Tan and
                  Shafiq R. Joty and
                  Enamul Hoque},
  title        = {ChartQA: {A} Benchmark for Question Answering about Charts with Visual
                  and Logical Reasoning},
  booktitle    = {Findings of the Association for Computational Linguistics: {ACL} 2022,
                  Dublin, Ireland, May 22-27, 2022},
  pages        = {2263--2279},
  publisher    = {Association for Computational Linguistics},
  year         = {2022},
}

@inproceedings{DBLP:conf/wacv/MathewKJ21,
  author       = {Minesh Mathew and
                  Dimosthenis Karatzas and
                  C. V. Jawahar},
  title        = {DocVQA: {A} Dataset for {VQA} on Document Images},
  booktitle    = {{IEEE} Winter Conference on Applications of Computer Vision, {WACV}
                  2021, Waikoloa, HI, USA, January 3-8, 2021},
  pages        = {2199--2208},
  publisher    = {{IEEE}},
  year         = {2021},
}

@inproceedings{DBLP:conf/wacv/MathewBTKVJ22,
  author       = {Minesh Mathew and
                  Viraj Bagal and
                  Rub{\`{e}}n Tito and
                  Dimosthenis Karatzas and
                  Ernest Valveny and
                  C. V. Jawahar},
  title        = {InfographicVQA},
  booktitle    = {{IEEE/CVF} Winter Conference on Applications of Computer Vision, {WACV}
                  2022, Waikoloa, HI, USA, January 3-8, 2022},
  pages        = {2582--2591},
  publisher    = {{IEEE}},
  year         = {2022},
}

@inproceedings{DBLP:conf/eccv/LiuDZLZZYWHLCL24,
  author       = {Yuan Liu and
                  Haodong Duan and
                  Yuanhan Zhang and
                  Bo Li and
                  Songyang Zhang and
                  Wangbo Zhao and
                  Yike Yuan and
                  Jiaqi Wang and
                  Conghui He and
                  Ziwei Liu and
                  Kai Chen and
                  Dahua Lin},
  title        = {MMBench: Is Your Multi-modal Model an All-Around Player?},
  booktitle    = {Computer Vision - {ECCV} 2024 - 18th European Conference, Milan, Italy,
                  September 29-October 4, 2024, Proceedings, Part {VI}},
  series       = {Lecture Notes in Computer Science},
  volume       = {15064},
  pages        = {216--233},
  publisher    = {Springer},
  year         = {2024},
}

@inproceedings{DBLP:conf/icml/YuYLWL0WW24,
  author       = {Weihao Yu and
                  Zhengyuan Yang and
                  Linjie Li and
                  Jianfeng Wang and
                  Kevin Lin and
                  Zicheng Liu and
                  Xinchao Wang and
                  Lijuan Wang},
  title        = {MM-Vet: Evaluating Large Multimodal Models for Integrated Capabilities},
  booktitle    = {Forty-first International Conference on Machine Learning, {ICML} 2024,
                  Vienna, Austria, July 21-27, 2024},
  publisher    = {OpenReview.net},
  year         = {2024},
}

@inproceedings{DBLP:conf/eccv/ZhangJZLGQZLCQGL24,
  author       = {Renrui Zhang and
                  Dongzhi Jiang and
                  Yichi Zhang and
                  Haokun Lin and
                  Ziyu Guo and
                  Pengshuo Qiu and
                  Aojun Zhou and
                  Pan Lu and
                  Kai{-}Wei Chang and
                  Yu Qiao and
                  Peng Gao and
                  Hongsheng Li},
  title        = {{MATHVERSE:} Does Your Multi-modal {LLM} Truly See the Diagrams in
                  Visual Math Problems?},
  booktitle    = {Computer Vision - {ECCV} 2024 - 18th European Conference, Milan, Italy,
                  September 29-October 4, 2024, Proceedings, Part {VIII}},
  series       = {Lecture Notes in Computer Science},
  volume       = {15066},
  pages        = {169--186},
  publisher    = {Springer},
  year         = {2024},
}

@article{DBLP:journals/corr/abs-2310-02255,
  author       = {Pan Lu and
                  Hritik Bansal and
                  Tony Xia and
                  Jiacheng Liu and
                  Chunyuan Li and
                  Hannaneh Hajishirzi and
                  Hao Cheng and
                  Kai{-}Wei Chang and
                  Michel Galley and
                  Jianfeng Gao},
  title        = {MathVista: Evaluating Math Reasoning in Visual Contexts with GPT-4V,
                  Bard, and Other Large Multimodal Models},
  journal      = {CoRR},
  volume       = {abs/2310.02255},
  year         = {2023},
}

@inproceedings{DBLP:conf/cvpr/YueNZ0LZSJRSWYY24,
  author       = {Xiang Yue and
                  Yuansheng Ni and
                  Tianyu Zheng and
                  Kai Zhang and
                  Ruoqi Liu and
                  Ge Zhang and
                  Samuel Stevens and
                  Dongfu Jiang and
                  Weiming Ren and
                  Yuxuan Sun and
                  Cong Wei and
                  Botao Yu and
                  Ruibin Yuan and
                  Renliang Sun and
                  Ming Yin and
                  Boyuan Zheng and
                  Zhenzhu Yang and
                  Yibo Liu and
                  Wenhao Huang and
                  Huan Sun and
                  Yu Su and
                  Wenhu Chen},
  title        = {{MMMU:} {A} Massive Multi-Discipline Multimodal Understanding and
                  Reasoning Benchmark for Expert {AGI}},
  booktitle    = {{IEEE/CVF} Conference on Computer Vision and Pattern Recognition,
                  {CVPR} 2024, Seattle, WA, USA, June 16-22, 2024},
  pages        = {9556--9567},
  publisher    = {{IEEE}},
  year         = {2024},
}

@inproceedings{DBLP:conf/naacl/ZhangLZPCHLZYLL25,
  author       = {Kaichen Zhang and
                  Bo Li and
                  Peiyuan Zhang and
                  Fanyi Pu and
                  Joshua Adrian Cahyono and
                  Kairui Hu and
                  Shuai Liu and
                  Yuanhan Zhang and
                  Jingkang Yang and
                  Chunyuan Li and
                  Ziwei Liu},
  title        = {LMMs-Eval: Reality Check on the Evaluation of Large Multimodal Models},
  booktitle    = {Findings of the Association for Computational Linguistics: {NAACL}
                  2025, Albuquerque, New Mexico, USA, April 29 - May 4, 2025},
  pages        = {881--916},
  publisher    = {Association for Computational Linguistics},
  year         = {2025},
}

@misc{Qwen3VL,
  author       = {{Qwen}},
  title        = {{Qwen3-VL}},
  howpublished = {\url{https://qwen.ai/}},
  year         = {2025},
}

@inproceedings{DBLP:conf/cvpr/LinYP0SH24,
  author       = {Ji Lin and
                  Hongxu Yin and
                  Wei Ping and
                  Pavlo Molchanov and
                  Mohammad Shoeybi and
                  Song Han},
  title        = {{VILA:} On Pre-training for Visual Language Models},
  booktitle    = {{IEEE/CVF} Conference on Computer Vision and Pattern Recognition,
                  {CVPR} 2024, Seattle, WA, USA, June 16-22, 2024},
  pages        = {26679--26689},
  publisher    = {{IEEE}},
  year         = {2024},
}

@article{DBLP:journals/corr/abs-2407-03320,
  author       = {Pan Zhang and
                  Xiaoyi Dong and
                  Yuhang Zang and
                  Yuhang Cao and
                  Rui Qian and
                  Lin Chen and
                  Qipeng Guo and
                  Haodong Duan and
                  Bin Wang and
                  Linke Ouyang and
                  Songyang Zhang and
                  Wenwei Zhang and
                  Yining Li and
                  Yang Gao and
                  Peng Sun and
                  Xinyue Zhang and
                  Wei Li and
                  Jingwen Li and
                  Wenhai Wang and
                  Hang Yan and
                  Conghui He and
                  Xingcheng Zhang and
                  Kai Chen and
                  Jifeng Dai and
                  Yu Qiao and
                  Dahua Lin and
                  Jiaqi Wang},
  title        = {InternLM-XComposer-2.5: {A} Versatile Large Vision Language Model
                  Supporting Long-Contextual Input and Output},
  journal      = {CoRR},
  volume       = {abs/2407.03320},
  year         = {2024},
}

@article{DBLP:journals/corr/abs-2312-14238,
  author       = {Zhe Chen and
                  Jiannan Wu and
                  Wenhai Wang and
                  Weijie Su and
                  Guo Chen and
                  Sen Xing and
                  Muyan Zhong and
                  Qinglong Zhang and
                  Xizhou Zhu and
                  Lewei Lu and
                  Bin Li and
                  Ping Luo and
                  Tong Lu and
                  Yu Qiao and
                  Jifeng Dai},
  title        = {InternVL: Scaling up Vision Foundation Models and Aligning for Generic
                  Visual-Linguistic Tasks},
  journal      = {CoRR},
  volume       = {abs/2312.14238},
  year         = {2023},
}

@inproceedings{DBLP:conf/nips/BrownMRSKDNSSAA20,
  author       = {Tom B. Brown and
                  Benjamin Mann and
                  Nick Ryder and
                  Melanie Subbiah and
                  Jared Kaplan and
                  Prafulla Dhariwal and
                  Arvind Neelakantan and
                  Pranav Shyam and
                  Girish Sastry and
                  Amanda Askell and
                  Sandhini Agarwal and
                  Ariel Herbert{-}Voss and
                  Gretchen Krueger and
                  Tom Henighan and
                  Rewon Child and
                  Aditya Ramesh and
                  Daniel M. Ziegler and
                  Jeffrey Wu and
                  Clemens Winter and
                  Christopher Hesse and
                  Mark Chen and
                  Eric Sigler and
                  Mateusz Litwin and
                  Scott Gray and
                  Benjamin Chess and
                  Jack Clark and
                  Christopher Berner and
                  Sam McCandlish and
                  Alec Radford and
                  Ilya Sutskever and
                  Dario Amodei},
  title        = {Language Models are Few-Shot Learners},
  booktitle    = {Advances in Neural Information Processing Systems 33: Annual Conference
                  on Neural Information Processing Systems 2020, NeurIPS 2020, December
                  6-12, 2020, virtual},
  year         = {2020},
}

@article{DBLP:journals/corr/abs-2302-13971,
  author       = {Hugo Touvron and
                  Thibaut Lavril and
                  Gautier Izacard and
                  Xavier Martinet and
                  Marie{-}Anne Lachaux and
                  Timoth{\'{e}}e Lacroix and
                  Baptiste Rozi{\`{e}}re and
                  Naman Goyal and
                  Eric Hambro and
                  Faisal Azhar and
                  Aur{\'{e}}lien Rodriguez and
                  Armand Joulin and
                  Edouard Grave and
                  Guillaume Lample},
  title        = {LLaMA: Open and Efficient Foundation Language Models},
  journal      = {CoRR},
  volume       = {abs/2302.13971},
  year         = {2023},
}

@inproceedings{DBLP:conf/nips/AlayracDLMBHLMM22,
  author       = {Jean{-}Baptiste Alayrac and
                  Jeff Donahue and
                  Pauline Luc and
                  Antoine Miech and
                  Iain Barr and
                  Yana Hasson and
                  Karel Lenc and
                  Arthur Mensch and
                  Katherine Millican and
                  Malcolm Reynolds and
                  Roman Ring and
                  Eliza Rutherford and
                  Serkan Cabi and
                  Tengda Han and
                  Zhitao Gong and
                  Sina Samangooei and
                  Marianne Monteiro and
                  Jacob L. Menick and
                  Sebastian Borgeaud and
                  Andy Brock and
                  Aida Nematzadeh and
                  Sahand Sharifzadeh and
                  Mikolaj Binkowski and
                  Ricardo Barreira and
                  Oriol Vinyals and
                  Andrew Zisserman and
                  Kar{\'{e}}n Simonyan},
  title        = {Flamingo: a Visual Language Model for Few-Shot Learning},
  booktitle    = {Advances in Neural Information Processing Systems 35: Annual Conference
                  on Neural Information Processing Systems 2022, NeurIPS 2022, New Orleans,
                  LA, USA, November 28 - December 9, 2022},
  year         = {2022},
}

@inproceedings{DBLP:conf/nips/LiuLWL23a,
  author       = {Haotian Liu and
                  Chunyuan Li and
                  Qingyang Wu and
                  Yong Jae Lee},
  title        = {Visual Instruction Tuning},
  booktitle    = {Advances in Neural Information Processing Systems 36: Annual Conference
                  on Neural Information Processing Systems 2023, NeurIPS 2023, New Orleans,
                  LA, USA, December 10 - 16, 2023},
  year         = {2023},
}

@inproceedings{DBLP:conf/nips/Dai0LTZW0FH23,
  author       = {Wenliang Dai and
                  Junnan Li and
                  Dongxu Li and
                  Anthony Meng Huat Tiong and
                  Junqi Zhao and
                  Weisheng Wang and
                  Boyang Li and
                  Pascale Fung and
                  Steven C. H. Hoi},
  title        = {InstructBLIP: Towards General-purpose Vision-Language Models with
                  Instruction Tuning},
  booktitle    = {Advances in Neural Information Processing Systems 36: Annual Conference
                  on Neural Information Processing Systems 2023, NeurIPS 2023, New Orleans,
                  LA, USA, December 10 - 16, 2023},
  year         = {2023},
}

@inproceedings{DBLP:conf/iclr/Zhu0SLE24,
  author       = {Deyao Zhu and
                  Jun Chen and
                  Xiaoqian Shen and
                  Xiang Li and
                  Mohamed Elhoseiny},
  title        = {MiniGPT-4: Enhancing Vision-Language Understanding with Advanced Large
                  Language Models},
  booktitle    = {The Twelfth International Conference on Learning Representations,
                  {ICLR} 2024, Vienna, Austria, May 7-11, 2024},
  publisher    = {OpenReview.net},
  year         = {2024},
}

@article{DBLP:journals/corr/abs-2310-09478,
  author       = {Jun Chen and
                  Deyao Zhu and
                  Xiaoqian Shen and
                  Xiang Li and
                  Zechun Liu and
                  Pengchuan Zhang and
                  Raghuraman Krishnamoorthi and
                  Vikas Chandra and
                  Yunyang Xiong and
                  Mohamed Elhoseiny},
  title        = {MiniGPT-v2: large language model as a unified interface for vision-language
                  multi-task learning},
  journal      = {CoRR},
  volume       = {abs/2310.09478},
  year         = {2023},
}

@article{DBLP:journals/corr/abs-2308-12966,
  author       = {Jinze Bai and
                  Shuai Bai and
                  Shusheng Yang and
                  Shijie Wang and
                  Sinan Tan and
                  Peng Wang and
                  Junyang Lin and
                  Chang Zhou and
                  Jingren Zhou},
  title        = {Qwen-VL: {A} Frontier Large Vision-Language Model with Versatile Abilities},
  journal      = {CoRR},
  volume       = {abs/2308.12966},
  year         = {2023},
}

@article{DBLP:journals/corr/abs-2412-05271,
  author       = {Zhe Chen and
                  Weiyun Wang and
                  Yue Cao and
                  Yangzhou Liu and
                  Zhangwei Gao and
                  Erfei Cui and
                  Jinguo Zhu and
                  Shenglong Ye and
                  Hao Tian and
                  Zhaoyang Liu and
                  Lixin Gu and
                  Xuehui Wang and
                  Qingyun Li and
                  Yimin Ren and
                  Zixuan Chen and
                  Jiapeng Luo and
                  Jiahao Wang and
                  Tan Jiang and
                  Bo Wang and
                  Conghui He and
                  Botian Shi and
                  Xingcheng Zhang and
                  Han Lv and
                  Yi Wang and
                  Wenqi Shao and
                  Pei Chu and
                  Zhongying Tu and
                  Tong He and
                  Zhiyong Wu and
                  Huipeng Deng and
                  Jiaye Ge and
                  Kai Chen and
                  Min Dou and
                  Lewei Lu and
                  Xizhou Zhu and
                  Tong Lu and
                  Dahua Lin and
                  Yu Qiao and
                  Jifeng Dai and
                  Wenhai Wang},
  title        = {Expanding Performance Boundaries of Open-Source Multimodal Models
                  with Model, Data, and Test-Time Scaling},
  journal      = {CoRR},
  volume       = {abs/2412.05271},
  year         = {2024},
}

@article{DBLP:journals/corr/abs-2407-14177,
  author       = {Kaibing Chen and
                  Dong Shen and
                  Hanwen Zhong and
                  Huasong Zhong and
                  Kui Xia and
                  Di Xu and
                  Wei Yuan and
                  Yifei Hu and
                  Bin Wen and
                  Tianke Zhang and
                  Changyi Liu and
                  Dewen Fan and
                  Huihui Xiao and
                  Jiahong Wu and
                  Fan Yang and
                  Size Li and
                  Di Zhang},
  title        = {{EVLM:} An Efficient Vision-Language Model for Visual Understanding},
  journal      = {CoRR},
  volume       = {abs/2407.14177},
  year         = {2024},
}

@article{DBLP:journals/corr/abs-2504-09925,
  author       = {Zheng Liu and
                  Mengjie Liu and
                  Jingzhou Chen and
                  Jingwei Xu and
                  Bin Cui and
                  Conghui He and
                  Wentao Zhang},
  title        = {{FUSION:} Fully Integration of Vision-Language Representations for
                  Deep Cross-Modal Understanding},
  journal      = {CoRR},
  volume       = {abs/2504.09925},
  year         = {2025},
}

@inproceedings{DBLP:conf/nips/MengYTDW0024,
  author       = {Lingchen Meng and
                  Jianwei Yang and
                  Rui Tian and
                  Xiyang Dai and
                  Zuxuan Wu and
                  Jianfeng Gao and
                  Yu{-}Gang Jiang},
  title        = {DeepStack: Deeply Stacking Visual Tokens is Surprisingly Simple and
                  Effective for LMMs},
  booktitle    = {Advances in Neural Information Processing Systems 38: Annual Conference
                  on Neural Information Processing Systems 2024, NeurIPS 2024, Vancouver,
                  BC, Canada, December 10 - 15, 2024},
  year         = {2024},
}

@misc{Grok-1.5,
  author       = {{X.AI Corp}},
  title        = {{Grok-1.5 vision preview: Connecting the digital and physical worlds with our first multimodal model}},
  howpublished = {\url{ https://x.ai/blog/grok-1.5v}},
  year         = {2024},
}

@inproceedings{DBLP:conf/iclr/HuSWALWWC22,
  author       = {Edward J. Hu and
                  Yelong Shen and
                  Phillip Wallis and
                  Zeyuan Allen{-}Zhu and
                  Yuanzhi Li and
                  Shean Wang and
                  Lu Wang and
                  Weizhu Chen},
  title        = {LoRA: Low-Rank Adaptation of Large Language Models},
  booktitle    = {The Tenth International Conference on Learning Representations, {ICLR}
                  2022, Virtual Event, April 25-29, 2022},
  publisher    = {OpenReview.net},
  year         = {2022},
}

@inproceedings{starace2023probing,
  author       = {Giulio Starace and
                  Konstantinos Papakostas and
                  Rochelle Choenni and
                  Apostolos Panagiotopoulos and
                  Matteo Rosati and
                  Alina Leidinger and
                  Ekaterina Shutova},
  title        = {Probing LLMs for Joint Encoding of Linguistic Categories},
  booktitle    = {Findings of the Association for Computational Linguistics: {EMNLP}
                  2023, Singapore, December 6-10, 2023},
  pages        = {7158--7179},
  publisher    = {Association for Computational Linguistics},
  year         = {2023},
}

@article{amir2021deep,
  author       = {Shir Amir and
                  Yossi Gandelsman and
                  Shai Bagon and
                  Tali Dekel},
  title        = {Deep ViT Features as Dense Visual Descriptors},
  journal      = {CoRR},
  volume       = {abs/2112.05814},
  year         = {2021},
}

@article{song2025demystifying,
  author       = {Xinyuan Song and
                  Keyu Wang and
                  Pengxiang Li and
                  Lu Yin and
                  Shiwei Liu},
  title        = {Demystifying the Roles of {LLM} Layers in Retrieval, Knowledge, and
                  Reasoning},
  journal      = {CoRR},
  volume       = {abs/2510.02091},
  year         = {2025},
}

@inproceedings{yem2025plug,
  author       = {Jiabo Ye and
                  Haiyang Xu and
                  Haowei Liu and
                  Anwen Hu and
                  Ming Yan and
                  Qi Qian and
                  Ji Zhang and
                  Fei Huang and
                  Jingren Zhou},
  title        = {mPLUG-Owl3: Towards Long Image-Sequence Understanding in Multi-Modal
                  Large Language Models},
  booktitle    = {The Thirteenth International Conference on Learning Representations,
                  {ICLR} 2025, Singapore, April 24-28, 2025},
  publisher    = {OpenReview.net},
  year         = {2025},
}

@inproceedings{DBLP:conf/cvpr/HudsonM19,
  author       = {Drew A. Hudson and
                  Christopher D. Manning},
  title        = {{GQA:} {A} New Dataset for Real-World Visual Reasoning and Compositional
                  Question Answering},
  booktitle    = {{IEEE} Conference on Computer Vision and Pattern Recognition, {CVPR}
                  2019, Long Beach, CA, USA, June 16-20, 2019},
  pages        = {6700--6709},
  publisher    = {Computer Vision Foundation / {IEEE}},
  year         = {2019},
}

@inproceedings{DBLP:conf/cvpr/MarinoRFM19,
  author       = {Kenneth Marino and
                  Mohammad Rastegari and
                  Ali Farhadi and
                  Roozbeh Mottaghi},
  title        = {{OK-VQA:} {A} Visual Question Answering Benchmark Requiring External
                  Knowledge},
  booktitle    = {{IEEE} Conference on Computer Vision and Pattern Recognition, {CVPR}
                  2019, Long Beach, CA, USA, June 16-20, 2019},
  pages        = {3195--3204},
  publisher    = {Computer Vision Foundation / {IEEE}},
  year         = {2019},
}

@inproceedings{DBLP:conf/nips/LuMX0CZTCK22,
  author       = {Pan Lu and
                  Swaroop Mishra and
                  Tanglin Xia and
                  Liang Qiu and
                  Kai{-}Wei Chang and
                  Song{-}Chun Zhu and
                  Oyvind Tafjord and
                  Peter Clark and
                  Ashwin Kalyan},
  title        = {Learn to Explain: Multimodal Reasoning via Thought Chains for Science
                  Question Answering},
  booktitle    = {Advances in Neural Information Processing Systems 35: Annual Conference
                  on Neural Information Processing Systems 2022, NeurIPS 2022, New Orleans,
                  LA, USA, November 28 - December 9, 2022},
  year         = {2022},
}

@inproceedings{DBLP:conf/acl/YingCWJWYSKYD25,
  author       = {Jie Ying and
                  Zihong Chen and
                  Zhefan Wang and
                  Wanli Jiang and
                  Chenyang Wang and
                  Zhonghang Yuan and
                  Haoyang Su and
                  Huanjun Kong and
                  Fan Yang and
                  Nanqing Dong},
  title        = {SeedBench: {A} Multi-task Benchmark for Evaluating Large Language
                  Models in Seed Science},
  booktitle    = {Proceedings of the 63rd Annual Meeting of the Association for Computational
                  Linguistics (Volume 1: Long Papers), {ACL} 2025, Vienna, Austria,
                  July 27 - August 1, 2025},
  pages        = {31395--31449},
  publisher    = {Association for Computational Linguistics},
  year         = {2025},
}

@inproceedings{DBLP:conf/nips/ChenLDZZCDWQLZ24,
  author       = {Lin Chen and
                  Jinsong Li and
                  Xiaoyi Dong and
                  Pan Zhang and
                  Yuhang Zang and
                  Zehui Chen and
                  Haodong Duan and
                  Jiaqi Wang and
                  Yu Qiao and
                  Dahua Lin and
                  Feng Zhao},
  title        = {Are We on the Right Way for Evaluating Large Vision-Language Models?},
  booktitle    = {Advances in Neural Information Processing Systems 38: Annual Conference
                  on Neural Information Processing Systems 2024, NeurIPS 2024, Vancouver,
                  BC, Canada, December 10 - 15, 2024},
  year         = {2024},
}

@inproceedings{DBLP:conf/emnlp/LiDZWZW23,
  author       = {Yifan Li and
                  Yifan Du and
                  Kun Zhou and
                  Jinpeng Wang and
                  Wayne Xin Zhao and
                  Ji{-}Rong Wen},
  title        = {Evaluating Object Hallucination in Large Vision-Language Models},
  booktitle    = {Proceedings of the 2023 Conference on Empirical Methods in Natural
                  Language Processing, {EMNLP} 2023, Singapore, December 6-10, 2023},
  pages        = {292--305},
  publisher    = {Association for Computational Linguistics},
  year         = {2023},
}

@article{DBLP:journals/corr/abs-2306-13394,
  author       = {Chaoyou Fu and
                  Peixian Chen and
                  Yunhang Shen and
                  Yulei Qin and
                  Mengdan Zhang and
                  Xu Lin and
                  Zhenyu Qiu and
                  Wei Lin and
                  Jinrui Yang and
                  Xiawu Zheng and
                  Ke Li and
                  Xing Sun and
                  Rongrong Ji},
  title        = {{MME:} {A} Comprehensive Evaluation Benchmark for Multimodal Large
                  Language Models},
  journal      = {CoRR},
  volume       = {abs/2306.13394},
  year         = {2023},
}

@article{wang2024cogvlm,
  title={Cogvlm: Visual expert for pretrained language models},
  author={Wang, Weihan and Lv, Qingsong and Yu, Wenmeng and Hong, Wenyi and Qi, Ji and Wang, Yan and Ji, Junhui and Yang, Zhuoyi and Zhao, Lei and XiXuan, Song and others},
  journal={Advances in Neural Information Processing Systems},
  volume={37},
  pages={121475--121499},
  year={2024}
}

@inproceedings{lin2025multi,
  title={Multi-layer visual feature fusion in multimodal llms: Methods, analysis, and best practices},
  author={Lin, Junyan and Chen, Haoran and Fan, Yue and Fan, Yingqi and Jin, Xin and Su, Hui and Fu, Jinlan and Shen, Xiaoyu},
  booktitle={Proceedings of the Computer Vision and Pattern Recognition Conference},
  pages={4156--4166},
  year={2025}
}

@article{wei2025dynamic,
  title={Dynamic Embedding of Hierarchical Visual Features for Efficient Vision-Language Fine-Tuning},
  author={Wei, Xinyu and Yang, Guoli and Zhou, Jialu and Yang, Mingyue and Li, Leqian and Zhang, Kedi and Qiu, Chunping},
  journal={arXiv preprint arXiv:2508.17638},
  year={2025}
}

@inproceedings{hong2024cogagent,
  title={Cogagent: A visual language model for gui agents},
  author={Hong, Wenyi and Wang, Weihan and Lv, Qingsong and Xu, Jiazheng and Yu, Wenmeng and Ji, Junhui and Wang, Yan and Wang, Zihan and Dong, Yuxiao and Ding, Ming and others},
  booktitle={Proceedings of the IEEE/CVF conference on computer vision and pattern recognition},
  pages={14281--14290},
  year={2024}
}
liji
\newpage
\appendix
\section{Model Configuration}
\label{app:config}
This section provides a detailed description of the model configurations and training protocols for the baseline models used in our experiments, namely LLaVA-OneVision and LLaVA-1.5.
The settings are adopted from their original papers \cite{DBLP:journals/tmlr/0080ZGZ00ZZL0L25,DBLP:conf/cvpr/LiuLLL24} to ensure a fair and reproducible comparison.

\subsection{LLaVA-OneVision}

\paragraph{Architecture.} 
Our methodology adopts the LLaVA-OneVision architecture \cite{DBLP:journals/tmlr/0080ZGZ00ZZL0L25} as its foundational structure. 
To investigate scalability, our experiments are conducted using its publicly available 0.5B and 7B parameter variants. 
This established VLM framework comprises three core components that enable its cross-modal capabilities. 

First, the language model (LLM) backbone consists of the Qwen2 series models, specifically Qwen2-0.5B and Qwen2-7B \cite{DBLP:journals/corr/abs-2407-10671}. The LLaVA-OneVision framework leverages these models for their robust language understanding and complex reasoning capabilities, which are essential for interpreting and acting upon user instructions. Second, visual understanding is handled by a pre-trained SigLIP vision encoder, Siglip-so400m-patch14-384 \cite{DBLP:conf/iccv/ZhaiM0B23}. 
The selection of this encoder provides the baseline with a powerful and generalizable foundation for feature extraction, stemming from its extensive pre-training on diverse web-scale data. 
Finally, a projector module, implemented as a two-layer Multilayer Perceptron (MLP), serves to bridge the two modalities. 
Its function within the architecture is to map the visual features from the SigLIP encoder's output space into the LLM's input embedding space, thereby allowing the language model to process visual and textual information in a unified manner.

\paragraph{CLI Configuration Details.}
For both the 0.5B and 7B variants of our CLI-enhanced LLaVA-OneVision models, we implemented a consistent high-density, many-to-many injection strategy. 
We extract hierarchical visual features from the 28-layer SigLIP vision encoder by sampling every \textbf{fourth} layer. 
These multi-level features are then injected into the LLM decoder at \textbf{corresponding intervals}. 
Specifically, for the LLaVA-OV-0.5B model, which comprises a 24-layer Qwen2-0.5B backbone, injections occur at every fourth layer. 
A similar strategy with a stride of four is applied to the 28-layer Qwen2-7B backbone of the LLaVA-OV-7B model.

To harmonize the features from diverse vision layers, our \textbf{Adaptive Multi-Projection (AMP)} module augments the pre-trained projector with layer-specific Low-Rank Adaptation (LoRA). 
For these adaptations, we configure the LoRA parameters with a rank of 128 and an alpha of 128.
At each designated injection point within the LLM, an \textbf{Adaptive Gating Fusion (AGF)} module governs the selective integration of these hierarchical visual features. 
The AGF mechanism first distills information from both modalities independently. 
For the visual input, a learnable query vector ($q_v$) attends to the projected visual features (acting as key and value) via multi-head self-attention. 
Concurrently, another learnable query ($q_h$) probes the LLM's current hidden states to produce a contextual summary.
The resulting context vectors from both attention modules are concatenated and processed by a gate controller, implemented as a linear layer. 
Finally, a Sigmoid activation function is applied to this fused representation to produce a dynamic weight, which determines the degree to which the new visual information updates the LLM's hidden state.

\paragraph{Training Strategy.} 
Our training diverges from the full curriculum of the original LLaVA-OneVision. 
We start from the official checkpoint released after its ``Stage-1.5: High-Quality Knowledge Learning" phase. Our training consists solely of the subsequent \textbf{Single-Image Fine-tuning} stage. 
This stage involves fine-tuning the entire model on an approximately 1.4 million sample dataset. 
This dataset was curated by randomly sampling from the full 3.2M single-image instruction data pool described in the original paper. 
The detailed composition of our sampled dataset is provided in Appendix \ref{app:data}. 
To select the optimal checkpoint and prevent overfitting, the training was conducted for up to three epochs, governed by an early stopping mechanism based on performance on a held-out validation set. The specific hyperparameters for our fine-tuning process are detailed in Tab. \ref{tab:config_onevision}.

\begin{table}[htbp]
\centering
\caption{Fine-tuning configuration for our LLaVA-OneVision models. This table details the settings for the single-image fine-tuning stage we performed.}
\label{tab:config_onevision}
\begin{tabular}{@{}lc@{}}
\toprule
\textbf{Parameter} & \textbf{Value} \\ \midrule
\multicolumn{2}{@{}l}{\textbf{Vision}} \\
\quad Resolution & $384 \times \{\{1 \times 1\}, \dots, \{6 \times 6\}\}$ \\
\quad Max \#Tokens & $729 \times 10$ \\ \midrule
\multicolumn{2}{@{}l}{\textbf{Data}} \\
\quad Dataset & Single-Image Instruction Data \\
\quad \#Samples & $\sim$1.4M (Sampled) \\ \midrule
\multicolumn{2}{@{}l}{\textbf{Model}} \\
\quad Trainable & Full Model \\
\quad Parameters (0.5B) & 0.8B \\
\quad Parameters (7B) & 8.0B \\ \midrule
\multicolumn{2}{@{}l}{\textbf{Training Hyperparameters}} \\
\quad Batch Size (0.5B) & 256 \\
\quad Batch Size (7B) & 256 \\
\quad LR: Vision Encoder & $2 \times 10^{-6}$ \\
\quad LR: Projector \& LLM & $1 \times 10^{-5}$ \\
\quad Epochs & Up to 3 (with Early Stopping) \\ \bottomrule
\end{tabular}
\end{table}

\subsection{LLaVA-1.5}

\paragraph{Architecture.} 
In parallel with our primary experiments, we also benchmark our approach against the widely-used LLaVA-1.5 framework \cite{DBLP:conf/cvpr/LiuLLL24}, specifically leveraging its 7B parameter version to ensure a fair comparison. 
This established architecture also integrates three distinct components to facilitate its cross-modal understanding. 
First, the core of its reasoning and generative capabilities is the Vicuna-7B (v1.5) model \cite{vicuna2023}, which serves as the language backbone. 
Second, for visual perception, the framework employs a pre-trained CLIP vision encoder (CLIP-ViT-L-336px) \cite{DBLP:conf/icml/RadfordKHRGASAM21}. 
This module is configured to process input images at a resolution of 336x336 pixels to extract high-level visual features. 
Finally, a two-layer Multilayer Perceptron (MLP), featuring a GELU activation function, acts as the projector. 
This crucial component connects the vision and language modalities by mapping the output features from the CLIP encoder into the input embedding space of the Vicuna language model.

\paragraph{CLI Implementation on LLaVA-1.5.}
To validate the architecture-agnostic nature of our framework, we integrated CLI into the LLaVA-1.5-7B model. 
This architecture features a different set of core components: a 32-layer Vicuna-7B model as the LLM backbone and a 24-layer CLIP-ViT-L-336px as the vision encoder. 
To maintain consistency with our primary experiments, we adopted the same high-density injection strategy. 
Specifically, we extract hierarchical visual features by sampling every \textbf{fourth} layer from the 24-layer CLIP encoder. 
These multi-level features are then injected into the 32-layer Vicuna decoder, also at a stride of every \textbf{fourth layer}. 
The core mechanisms of our CLI framework, including the LoRA-based \textbf{Adaptive Multi-Projection (AMP)} and the query-based \textbf{Adaptive Gating Fusion (AGF)} modules, remain identical to the configuration used for the LLaVA-OneVision models, ensuring a fair and direct comparison of the framework's generalizability.

\paragraph{Training Strategy.} 
Our LLaVA-1.5 experiments start from the official pre-trained checkpoint, which has completed Stage-1 (Vision-Language Alignment). 
Our work thus consists exclusively of the second stage: \textbf{Visual Instruction Tuning (Fine-tuning)}. 
To ensure a fair comparison between model architectures, we utilized the identical training dataset and procedure as in our LLaVA-OneVision experiments. Specifically, the entire model was fine-tuned on the same $\sim$1.4 million sampled single-image instruction dataset.
The hyperparameters for our fine-tuning process are listed in Tab. \ref{tab:config_llava1.5}.

\begin{table}[htbp]
\centering
\caption{Fine-tuning configuration for our LLaVA-1.5 model. This table details the settings for the visual instruction tuning stage we performed.}
\label{tab:config_llava1.5}
\begin{tabular}{@{}lc@{}}
\toprule
\textbf{Hyperparameter} & \textbf{Value} \\ \midrule
Batch Size              & 256                             \\
Learning Rate (LR)      & $2 \times 10^{-5}$              \\
LR Schedule             & Cosine Decay                    \\
LR Warmup Ratio         & 0.03                            \\
Weight Decay            & 0                               \\
Epochs                  & 1                               \\
Optimizer               & AdamW                           \\ \bottomrule
\end{tabular}
\end{table}

\section{Instruction Tuning Dataset Details}
\label{app:data}

Our fine-tuning process utilizes a meticulously curated instruction-following dataset of approximately 1.4 million samples. 
This dataset is a sampled subset of the comprehensive 3.2 million single-image data pool introduced in the LLaVA-OneVision paper \cite{DBLP:journals/tmlr/0080ZGZ00ZZL0L25}. The goal of our curation was to create a high-quality, balanced mixture that covers a wide range of visual tasks while remaining computationally efficient for our experiments.

\paragraph{Data Composition.}
Following the categorization of LLaVA-OneVision, our dataset is composed of five principal categories to ensure a diverse skill set for the fine-tuned models:
\begin{itemize}
    \item \textbf{General QA and Conversation:} This is the largest category, designed to enhance the model's core visual dialogue and question-answering capabilities. It includes data from sources like ShareGPT4V/4o, Vision FLAN, and the general-purpose instruction sets from Cambrian and ALLaVA.
    \item \textbf{Math and Reasoning:} To bolster the model's logical and spatial reasoning abilities, we incorporated a significant portion of math-related visual question-answering data. Key sources include Geo170K, MathQA, and various datasets from the MathV360K collection.
    \item \textbf{Document, Chart, and Screen Understanding:} This category focuses on fine-grained perception of structured information. We sampled from datasets such as UReader, ChartQA, AI2D, and InfographicVQA.
    \item \textbf{General OCR:} To improve text recognition in natural scenes, we included data from sources like TextCaps and IAM.
    \item \textbf{Language:} To maintain and enhance the underlying language capabilities of the LLM, we included a substantial amount of high-quality, text-only instruction data from the Magpie-Pro collection.
\end{itemize}

\paragraph{Sampling Strategy.}
Instead of using the entire 3.2M sample pool, we employed a curated sampling strategy to construct our final dataset. 
As detailed in Tab. \ref{tab:dataset_composition}, this involved taking all samples from some smaller, high-quality sources (e.g., GEOS, Diagram Image2Text) while applying percentage-based sampling to larger datasets. 
This approach allowed us to tailor the contribution of each data source, creating a balanced and effective training mixture. 
All data was formatted according to the LLaVA prompting strategy to ensure compatibility and avoid instructional conflicts.

\definecolor{category_bg}{rgb}{0.95,0.7,0.3} 
\begin{table*}[htbp]
\centering
\caption{Detailed composition of the $\sim$1.4M instruction tuning dataset used in our experiments.}
\label{tab:dataset_composition}
\small
\begin{tabularx}{\textwidth}{@{}X r | X r@{}}
\multicolumn{4}{c}{\cellcolor{category_bg}\textbf{General (1,011,544 samples)}} \\
\toprule
\textbf{Dataset} & \textbf{\# Samples} & \textbf{Dataset} & \textbf{\# Samples} \\
\midrule
ALLaVA Instruct & 20,994 & ScienceQA & 498 \\
AOKVQA & 1,654 & ShareGPT4o & 57,284 \\
Cambrian (filtered) & 83,125 & ShareGPT4V & 90,985 \\
Hateful Memes & 850 & TallyQA & 4,934 \\
IconQA & 1,130 & Vision FLAN (filtered) & 184,173 \\
Image Textualization & 9,958 & Visual7W & 1,437 \\
LLaVA-NeXT Base & 369,294 & VisualWebInstruct & 263,583 \\
LLaVAR & 1,979 & VisText & 997 \\
LRV Normal (filtered) & 1,049 & VizWiz & 661 \\
PMC-VQA & 360 & VQARAD & 31 \\
 & & VSR & 216 \\
 & & WebSight & 1,000 \\
\midrule[\heavyrulewidth]
\multicolumn{4}{c}{\cellcolor{category_bg}\textbf{Math/Reasoning (71,321 samples)}} \\
\toprule
\textbf{Dataset} & \textbf{\# Samples} & \textbf{Dataset} & \textbf{\# Samples} \\
\midrule
CLEVR-Math & 528 & MAVIS (Metagen \& Rule) & 18,734 \\
FigureQA & 1,759 & MathQA & 29,827 \\
GEOS & 498 & MapQA & 4,265 \\
Geo170K (Align \& QA) & 12,808 & Super-CLEVR & 865 \\
GeoMVerse & 930 & TabMWP & 3,382 \\
GeoQA+ & 1,717 & TQA & 1,366 \\
Geometry3K & 3,064 & UniGeo & 1,195 \\
InterGPS & 128 & & \\
\midrule[\heavyrulewidth]
\multicolumn{4}{c}{\cellcolor{category_bg}\textbf{Doc/Chart/Screen (77,349 samples)}} \\
\toprule
\textbf{Dataset} & \textbf{\# Samples} & \textbf{Dataset} & \textbf{\# Samples} \\
\midrule
AI2D & 8,534 & MultiHierTT (Cauldron) & 762 \\
Chart2Text & 2,696 & RoBUT SQA & 851 \\
ChartQA & 1,826 & RoBUT WikiSQL & 7,499 \\
Diagram Image2Text & 295 & Screen2Words & 1,573 \\
HiTab & 250 & UReader (Cap, IE, KG, QA) & 39,927 \\
Infographic VQA & 6,588 & VisualMRC & 303 \\
LRV Chart & 1,776 & & \\
\midrule[\heavyrulewidth]
\multicolumn{4}{c}{\cellcolor{category_bg}\textbf{General OCR (6,472 samples) \qquad \qquad Language (179,993 samples)}} \\
\toprule
\textbf{Dataset} & \textbf{\# Samples} & \textbf{Dataset} & \textbf{\# Samples} \\
\midrule
IAM & 566 & Magpie Pro (L3 MT) & 59,998 \\
Rendered Text & 1,000 & Magpie Pro (L3 ST) & 59,998 \\
ST-VQA & 1,725 & Magpie Pro (Qwen2 ST) & 59,997 \\
TextCaps & 2,195 & & \\
TextOCR (GPT4V) & 2,511 & & \\
\bottomrule
\end{tabularx}
\end{table*}

\section{Evaluation Benchmarks}
\label{app:benchmark}

To ensure a standardized and reproducible comparison, we evaluate all models across a comprehensive suite of single-image benchmarks using the open-source LMMs-Eval framework \cite{DBLP:conf/naacl/ZhangLZPCHLZYLL25}. 
These benchmarks are grouped into three primary categories to assess a wide range of capabilities, from fine-grained perception to complex, real-world reasoning. 
All evaluations are conducted in a zero-shot setting unless otherwise specified.

\paragraph{Evaluation Metrics.}
Our evaluation primarily relies on the standard metrics implemented within the LMMs-Eval framework. For most benchmarks involving definite answers (e.g., multiple-choice or single-word VQA), we report standard \textbf{Accuracy}. For text-centric VQA tasks like DocVQA, we use \textbf{Average Normalized Levenshtein Similarity (ANLS)} to robustly measure performance against potential OCR inaccuracies. For open-ended conversational benchmarks such as LLaVA-in-the-Wild, we employ \textbf{GPT-assisted evaluation}, where GPT-4 serves as a judge to score model responses.

We employ specific aggregation methods for two benchmarks to derive a single, comparable score:
\begin{itemize}
\item For the \textbf{MME} benchmark, we report a unified success rate. This is calculated by summing the raw \textit{cognition\_score} and \textit{perception\_score} and then dividing by the total number of samples across both categories.
\item For the \textbf{MathVerse} benchmark, the final score is the average accuracy across its three sub-tasks: \textit{vision\_intensive}, \textit{vision\_only}, and \textit{\textbf{}vision\_dominant}.
\end{itemize}

\begin{table*}[]
\caption{
To validate the architecture-agnostic nature of our method, we integrated CLI into the LLaVA-1.5 architecture. 
The CLI-enhanced model demonstrates consistent and broad performance gains across a wide range of tasks, from document understanding and real-world chat (top) to complex, multidisciplinary reasoning (bottom), confirming the robustness and general applicability of our framework.
}
\label{tab:app_llava15}
\renewcommand\arraystretch{1.2}
\renewcommand\tabcolsep{4.0pt}
\centering
\resizebox{\linewidth}{!}{
\begin{tabular}{c|ccccccccc|c}
\toprule
\textbf{Model} & AI2D & ChartQA & DocVQA & InfoVQA & RealWorldQA & LLaVA-W & POPE & OK-VQA & GQA & \textbf{Partial Sum} \\
\midrule
LLaVA-1.5-7B & \textbf{66.3} & 38.9 & 32.2/- & \textbf{26.8}/- & 54.5 & 62.9 & \textbf{87.0} & 41.8 & \textbf{57.6} & 468.0 \\
\rowcolor{mygray}\ \textit{w}/ DeepStack \cite{DBLP:conf/nips/MengYTDW0024} & 65.4\textsubscript{\color{mygreen}-0.9} & \textbf{41.4}\textsubscript{\color{myred}+2.5} & \textbf{33.1}\textsubscript{\color{myred}+0.9}/- & 26.2\textsubscript{\color{mygreen}-0.6}/- & 48.7\textsubscript{\color{mygreen}-5.8} & 47.7\textsubscript{\color{mygreen}-15.2} & 86.7\textsubscript{\color{mygreen}-0.3} & 43.3\textsubscript{\color{myred}+1.5} & 57.0\textsubscript{\color{mygreen}-0.6} & 449.5\textsubscript{\color{mygreen}-18.5} \\
\rowcolor{mygray}\ \textit{w}/ SLI \cite{qwen3vl} & 64.8\textsubscript{\color{mygreen}-1.5} & 38.0\textsubscript{\color{mygreen}-0.9} & 30.4\textsubscript{\color{mygreen}-1.8}/- & 24.7\textsubscript{\color{mygreen}-2.1}/- & 52.9\textsubscript{\color{mygreen}-1.6} & 56.1\textsubscript{\color{mygreen}-6.8} & 86.3\textsubscript{\color{mygreen}-0.7} & \textbf{50.6}\textsubscript{\color{myred}+8.8} & 57.0\textsubscript{\color{mygreen}-0.6} & 460.8\textsubscript{\color{mygreen}-7.2} \\
\rowcolor{mygray}
\ \textit{w}/ \textbf{CLI} & 65.7\textsubscript{\color{mygreen}-0.6} & 39.6\textsubscript{\color{myred}+0.7} & 32.4\textsubscript{\color{myred}+0.2}/- & 26.3\textsubscript{\color{mygreen}-0.5}/- & \textbf{54.6}\textsubscript{\color{myred}+0.1} & \textbf{65.9}\textsubscript{\color{myred}+3.0} & 86.4\textsubscript{\color{mygreen}-0.6} & 47.0\textsubscript{\color{myred}+5.2} & \textbf{57.6} & \textbf{475.5}\textsubscript{\color{myred}+7.5} \\
\midrule

\textbf{Model} & MathVerse & MathVista & MMBench & MME & MMStar & MMMU & MMVet & SeedBench & ScienceQA & \textbf{Partial Sum} \\
\midrule
LLaVA-1.5-7B & 17.5 & 34.4 & 64.3 & \textbf{79.9} & 37.2 & 34.5 & 31.2 & \textbf{61.9} & \textbf{72.9} & 433.8 \\
\rowcolor{mygray}\ \textit{w}/ DeepStack \cite{DBLP:conf/nips/MengYTDW0024} & 12.3\textsubscript{\color{mygreen}-5.2} & 34.8\textsubscript{\color{myred}+0.4} & 63.2\textsubscript{\color{mygreen}-1.1} & 76.0\textsubscript{\color{mygreen}-3.9} & 37.5\textsubscript{\color{myred}+0.3} & \textbf{36.5}\textsubscript{\color{myred}+2.0} & \textbf{34.5}\textsubscript{\color{myred}+3.3} & 61.0\textsubscript{\color{mygreen}-0.9} & 70.5\textsubscript{\color{mygreen}-2.4} & 426.3\textsubscript{\color{mygreen}-7.5} \\
\rowcolor{mygray}\ \textit{w}/ SLI \cite{qwen3vl} & 17.7\textsubscript{\color{myred}+0.2} & 33.7\textsubscript{\color{mygreen}-0.7} & 62.6\textsubscript{\color{mygreen}-1.7} & 76.6\textsubscript{\color{mygreen}-3.3} & 35.3\textsubscript{\color{mygreen}-1.9} & 35.3\textsubscript{\color{myred}+0.8} & 30.6\textsubscript{\color{mygreen}-0.6} & 51.6\textsubscript{\color{mygreen}-10.3} & 72.8\textsubscript{\color{mygreen}-0.1} & 416.2\textsubscript{\color{mygreen}-17.6} \\
\rowcolor{mygray}
\ \textit{w}/ \textbf{CLI} & \textbf{18.3}\textsubscript{\color{myred}+0.8} & \textbf{35.6}\textsubscript{\color{myred}+1.2} & \textbf{66.3}\textsubscript{\color{myred}+2.0} & 79.4\textsubscript{\color{mygreen}-0.5} & \textbf{38.5}\textsubscript{\color{myred}+1.3} & 35.7\textsubscript{\color{myred}+1.2} & \textbf{34.5}\textsubscript{\color{myred}+3.3} & \textbf{61.9} & 72.2\textsubscript{\color{mygreen}-0.7} & \textbf{442.4}\textsubscript{\color{myred}+8.6} \\
\bottomrule
\end{tabular}

}
\end{table*}

\paragraph{Chart, Diagram, and Document Understanding.}
This category of benchmarks tests the models' ability to perform fine-grained perception and reasoning on structured visual information, which often involves understanding text, layouts, and data visualizations.
\begin{itemize}
\item \textbf{AI2D} \cite{DBLP:conf/eccv/KembhaviSKSHF16}: A benchmark for question answering on science diagrams, requiring parsing of diagrammatic elements and text.
\item \textbf{ChartQA} \cite{DBLP:conf/acl/MasryLTJH22}: A question-answering dataset focused on charts, which demands both visual perception and logical reasoning over the data presented.
\item \textbf{DocVQA} \cite{DBLP:conf/wacv/MathewKJ21}: A dataset for visual question answering on document images, testing the model's ability to read and comprehend text in complex layouts.
\item \textbf{InfoVQA} \cite{DBLP:conf/wacv/MathewBTKVJ22}: A benchmark for VQA on infographics, requiring the model to synthesize information from a mix of text, charts, and images.
\end{itemize}

\paragraph{Perception and Multidisciplinary Reasoning.}
This group of benchmarks evaluates the models' core visual perception skills and their ability to integrate this perception with world knowledge for complex reasoning across multiple disciplines.
\begin{itemize}
\item \textbf{MME} \cite{DBLP:journals/corr/abs-2306-13394}: A comprehensive benchmark designed to evaluate both the perception and cognition capabilities of multimodal models.
\item \textbf{MMBench} \cite{DBLP:conf/eccv/LiuDZLZZYWHLCL24}: A multi-discipline benchmark that assesses models on a wide range of skills through multiple-choice questions.
\item \textbf{MMVet} \cite{DBLP:conf/icml/YuYLWL0WW24}: Evaluates large multimodal models for their integrated capabilities across six core vision-language domains.
\item \textbf{MathVerse} \cite{DBLP:conf/eccv/ZhangJZLGQZLCQGL24}: A benchmark focused on visual math problems, specifically testing the ability to understand diagrams and figures in a mathematical context.
\item \textbf{MathVista} \cite{DBLP:journals/corr/abs-2310-02255}: A comprehensive benchmark for evaluating mathematical reasoning in diverse visual contexts.
\item \textbf{MMMU} \cite{DBLP:conf/cvpr/YueNZ0LZSJRSWYY24}: A massive multi-discipline benchmark that assesses models on college-level problems requiring expert-level understanding and reasoning.
\item \textbf{GQA} \cite{DBLP:conf/cvpr/HudsonM19}: A dataset for real-world visual reasoning and compositional question answering.
\item \textbf{OK-VQA} \cite{DBLP:conf/cvpr/MarinoRFM19}: A visual question answering benchmark where questions require external knowledge beyond what is present in the image.
\item \textbf{ScienceQA} \cite{DBLP:conf/nips/LuMX0CZTCK22}: A multimodal dataset for science question answering, involving text, diagrams, and formulas.
\item \textbf{SEED-Bench} \cite{DBLP:conf/acl/YingCWJWYSKYD25}: A multi-task benchmark for evaluating the generative comprehension of LMMs.
\item \textbf{MM-Star} \cite{DBLP:conf/nips/ChenLDZZCDWQLZ24}: A benchmark designed with challenging examples to test the limits of advanced VLMs.
\item \textbf{POPE} \cite{DBLP:conf/emnlp/LiDZWZW23}: A benchmark specifically designed to evaluate object hallucination in large vision-language models by testing their polling-based object presence evaluation.
\end{itemize}

\paragraph{Real-world Understanding and Visual Chat.}
This category assesses the practical utility of models as general-purpose visual assistants in open-ended, real-world scenarios.
\begin{itemize}
\item \textbf{RealWorldQA} \cite{Grok-1.5}: A benchmark containing questions about real-world images, often requiring common-sense reasoning and detailed observation.
\item \textbf{LLaVA-in-the-Wild} \cite{DBLP:conf/cvpr/LiuLLL24}: A set of challenging, open-ended visual conversation prompts designed to evaluate the practical chat capabilities of LMMs in unconstrained scenarios.
\end{itemize}

\section{Deepstack and SLI in LLaVA-1.5}
\label{app:llava15_deepstack}

To validate the architecture-agnostic nature of our framework and conduct a rigorous comparison against alternative deep fusion strategies, we integrated \textbf{CLI}, \textbf{DeepStack}, and \textbf{SLI} into the LLaVA-1.5-7B architecture. The results, detailed in Tab.~\ref{tab:app_llava15}, not only confirm the general applicability of CLI but also highlight the fundamental limitations of non-adaptive fusion methods.

The brute-force, one-to-many injection of visual features of DeepStack proves highly detrimental, leading to a substantial performance degradation, with partial sum scores dropping by 18.5 and 7.5 points. 
This negative impact was particularly pronounced on conversational benchmarks like LLaVA-W ($-15.2\%$), supporting the hypothesis that unfiltered, non-selective feature injection disrupts the LLM's learned representations and generative coherence.

Similarly, the statically-wired, one-to-one approach of SLI resulted in a significant performance collapse. 
This method, which creates a rigid mapping between initial vision encoder layers and shallow LLM decoder layers, was particularly damaging for complex reasoning tasks. 
The partial sum score on multidisciplinary reasoning benchmarks plummeted by a catastrophic 17.6 points, with severe losses on benchmarks like LLaVA-W ($-6.8\%$) and SEED-Bench ($-10.3\%$). 
This failure demonstrates that confining multi-level visual information to the LLM's shallowest layers perceptually impoverishes the deeper, reasoning-intensive layers, leaving them unable to access the visual details required for sophisticated problem-solving.

In stark contrast, our \textbf{CLI} framework delivered consistent and significant performance improvements, boosting the partial sum scores by +7.5\% and +8.6\%. 
The dynamic, context-aware selection enabled by \textbf{CLI's} AGF module allows the LLM to selectively integrate the most relevant visual features, thereby enriching its understanding without disrupting its internal representations. 
This experiment thus validates \textbf{CLI} as a robust and superior paradigm for deep vision-language fusion whose benefits generalize across diverse model architectures.

\section{More Details on Ablation Studies}
\label{app:ablation}

To rigorously validate the design of our CLI framework, we conduct a series of ablation studies on LLaVA-OV-0.5B. 
These experiments are designed to analyze the impact of data scale, dissect the individual contributions of our key components, and determine the optimal injection strategy.

\begin{figure*}[t]
\includegraphics[width=1.0\linewidth]{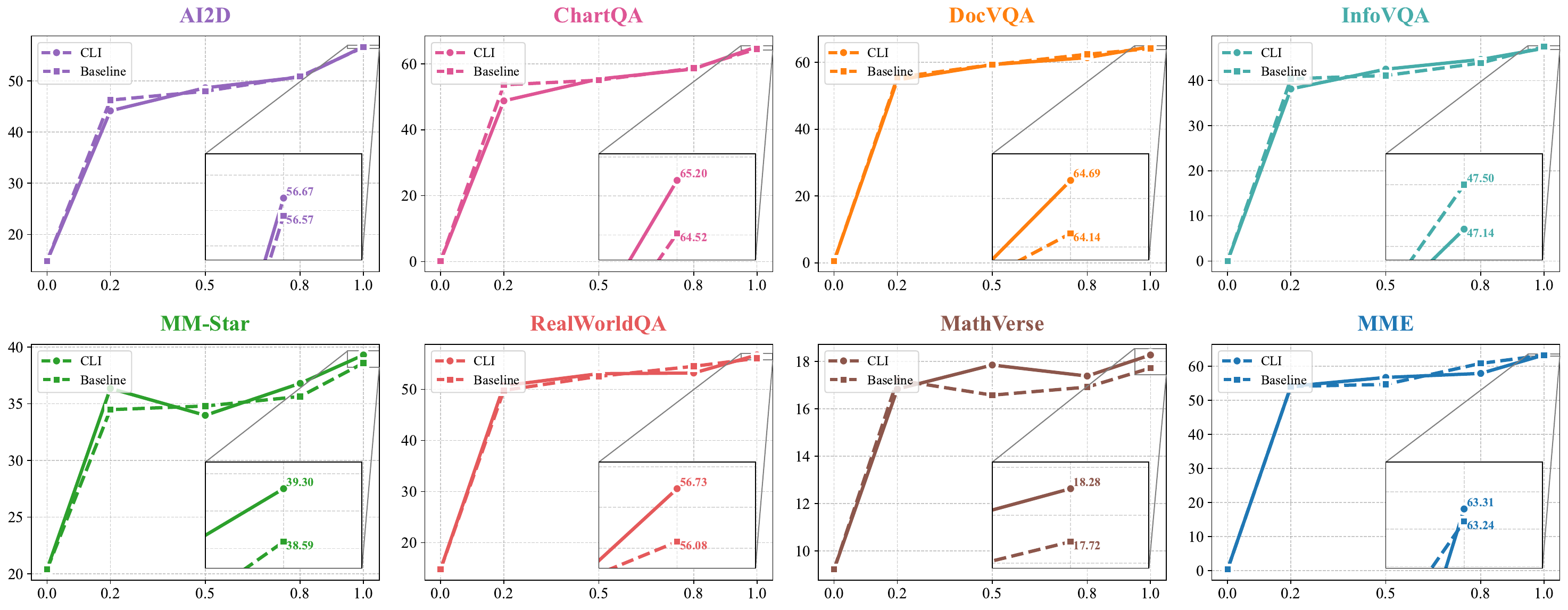}
\caption{Ablation study on the effect of training data volume. 
Performance of our proposed framework (CLI) and the baseline model is evaluated on eight benchmarks using varying percentages of the instruction tuning dataset. 
Our framework consistently improves performance and effectively leverages larger instruction sets, confirming its robustness at different data scales.}
\label{fig:data_volume}
\end{figure*}

\subsection{Impact of Instruction Data Volume}
\label{app:data_size}
To investigate the data efficiency and scalability of our CLI framework, we first analyze its performance when trained on varying subsets of the instruction tuning data (20\%, 50\%, and 80\%). 
As shown in Fig.~\ref{fig:data_volume}, the results reveal a nuanced relationship between data volume, task type, yielding five key insights.
\uline{First}, in the absence of any instruction tuning (the 0\% case), the model's performance is negligible, confirming the critical role of this training phase for instruction-following. 
\uline{Second}, performance gains are most significant in the low-data regime (from 0\% to 50\%), after which the rate of improvement begins to plateau, indicating diminishing returns. 
\uline{Third}, despite these diminishing returns, the CLI framework consistently achieves superior peak performance. 
At the 100\% data point, CLI outperforms or matches the baseline on seven of the eight benchmarks, demonstrating that our architecture is not only more data-efficient but also attains a higher final performance level. 
\uline{Fourth}, the amount of data required for CLI to surpass the baseline is highly task-dependent. 
For \textbf{fine-grained perception tasks} (e.g., DocVQA, InfoVQA) that necessitate precise OCR and layout understanding, CLI requires more data to realize its full advantage; its performance curve typically intersects and then surpasses the baseline's around the 50\% data mark, suggesting that learning to leverage multi-level details for high-fidelity perception is a data-intensive process. 
\uline{Fifth}, in contrast, for \textbf{reasoning-intensive tasks} (e.g., MM-Star, MathVerse), the architectural advantages of CLI are apparent even at low data volumes. 
The framework establishes a significant performance lead over the baseline with as little as 20\% of the training data, indicating that its dynamic integration of global context with local details provides a more data-efficient strategy for complex reasoning, enabling the model to acquire effective problem-solving heuristics more rapidly.
Based on this analysis, we identify the 50\% data subset as an efficient trade-off point for conducting further, more computationally intensive ablations.

\subsection{Dissecting the Contributions of CLI Components}
The efficacy of our framework is predicated upon two synergistic modules: the AMP for harmonizing hierarchical features and the AGF for their selective injection. 
To isolate their individual and combined impacts, we conducted a component-wise ablation study, with results presented in Tab.~\ref{ablation_result_1}.
We first test the effect of injecting multi-level features without a gating mechanism. Variants \textit{w/} AMP show only marginal gains over the baseline. 
This is a critical finding: merely flooding the LLM with hierarchical visual information is ineffective and risks causing information overload. 
Without a mechanism to filter for relevance, the model cannot effectively utilize the richer data stream.
Further, in stark contrast, a variant that incorporates only the AGF module yields a significant performance uplift. 
This result decisively validates our core hypothesis: the most crucial element for effective hierarchical fusion is empowering the LLM to act as an active observer, dynamically selecting visual information based on its real-time decoding context. 
The ability to selectively gate information is more important than the richness of the information itself.
Finally, we observe the synergy between our modules. 
When AMP is added to the gating mechanism, performance improves further. 
This confirms that adapting the projector to the distinct statistical distributions of different vision layers provides the AGF with a cleaner, more harmonized set of features, making its selection process more effective.

\begin{figure*}[tp]
\includegraphics[width=1.0\textwidth]{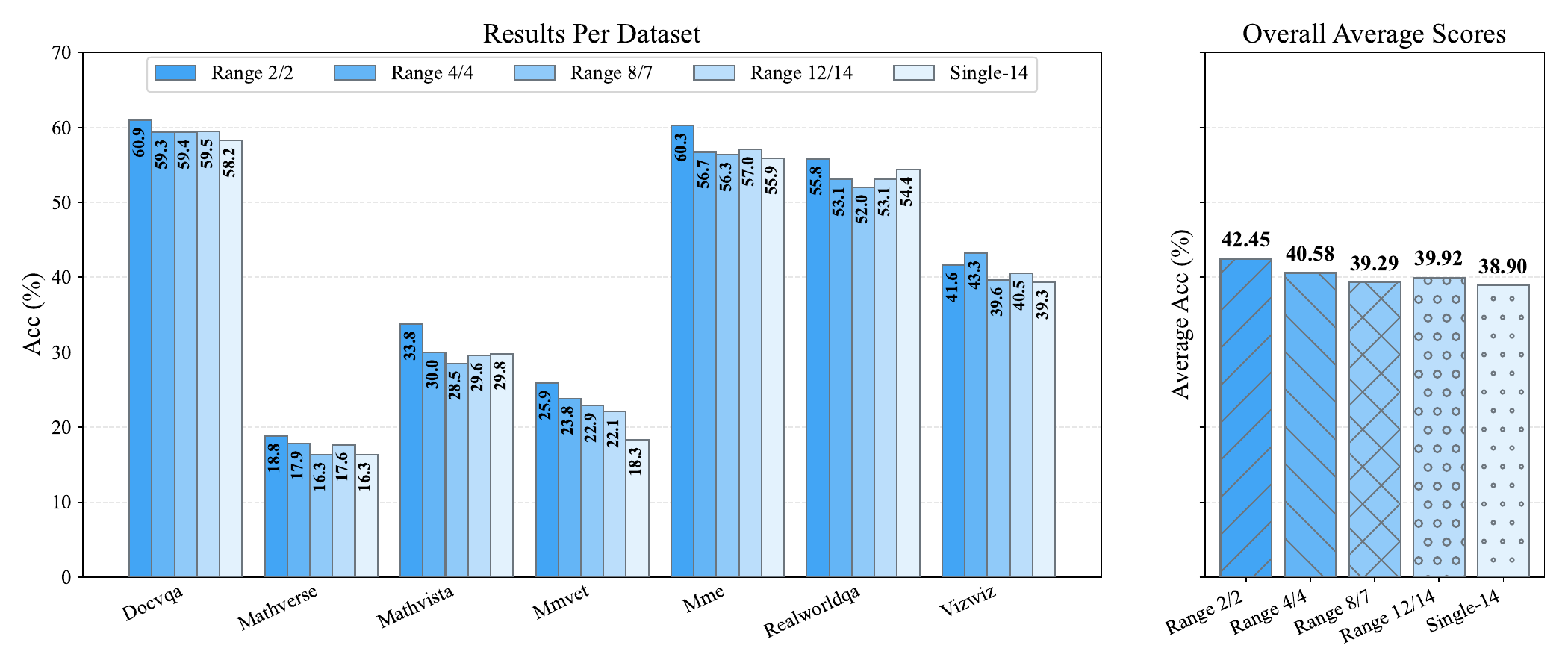}
\caption{Ablation study on the density of cross-layer injection points. 
The performance of different injection strategies—from high-density (`2/2') to a single-layer baseline (`Single-14')—is evaluated across several benchmarks. 
The high-density configuration achieves the best overall performance, validating our many-to-many design.}
\label{fig:layer_selection}
\end{figure*}

To further evaluate the efficiency of our proposed AMP, we conduct experiments where a dedicated projector is fully fine-tuned for each selected layer. 
As indicated in Tab.~\ref{ablation_result_1}, this full fine-tuning approach, despite its increased parameter count (107.95\% vs 102.25\%), results in performance degradation compared to our standard AMP (367.74 vs 367.89 ). 
We hypothesize this degradation stems from two key risks of full fine-tuning: catastrophic forgetting, by overwriting the projector's valuable pre-trained knowledge, and overfitting on the limited dataset due to the model's expanded parameter space.
However, when this fully fine-tuned projector is augmented with our AGF, it achieves the highest performance. 
A potential explanation is that while full fine-tuning unlocks maximum representational capacity, only a mechanism like AGF can effectively harness it by selecting salient features and mitigating noise.
Finally, given the superior balance between performance and parameter efficiency, we adopt the parameter-efficient LoRA-based AMP in our final framework, as it provides a compelling and scalable alternative to full fine-tuning.

\subsection{Impact of Injection Density}
Further, we conduct an ablation study by varying the density of cross-layer injection points to probe the effectiveness of our dynamic ``many-to-many'' architecture. 
The results, as shown in Fig.~\ref{fig:layer_selection}, are highly revealing: the `Single-14' baseline consistently underperforms, reaffirming that a single static feature map creates an insurmountable information bottleneck. 
Conversely, the high-density `2/2' strategy achieves the best overall performance, powerfully validating our principle that frequent access to the full visual hierarchy allows the AGF module to effectively navigate the dense information stream. 
Intriguingly, the study also uncovers a crucial trade-off between information gain and cognitive overhead, as the medium-density `8/7' configuration underperforms the sparser `12/14' strategy. 
We hypothesize that the `8/7' setting introduces disruptive cognitive overhead without the benefit of near-continuous context, whereas the `12/14' strategy offers a more ``economical'' and stable, albeit less powerful, update schedule.

\begin{table}[t]
\caption{Performance Impact of AGF's Query Source. 
This table evaluates the performance impact of the query source used by the AGF module. 
It compares two configurations: one where the gate controller is conditioned on visual tokens from the LLM's hidden state (Visual-Query, our method), and another conditioned on text tokens (Text-Query).}
\label{ablation_result_query}
\renewcommand\arraystretch{1.2}
\renewcommand\tabcolsep{4.0pt}
\centering
\resizebox{\linewidth}{!}{
\begin{tabular}{l|ccccccccc|c}
\toprule
\multicolumn{1}{c|}{\multirow{2}{*}{\textbf{Variant}}} & \textbf{MME} & \textbf{MM-Star} & \textbf{MMMU} & \textbf{SEED-Bench} & \textbf{AI2D} & \textbf{ChartQA} & \textbf{DocVQA} & \textbf{InfoVQA} & \textbf{RealWorldQA} & \multirow{2}{*}{\textbf{Sum}} \\ 
\cmidrule(lr){2-10}
        & test & test & val & image & test & test & val & val & test & \\ \midrule 
Baseline & 85.11 & 54.16 & 47.00 & 76.18 & 77.56 & 78.52 & 82.54 & 69.54 & \textbf{68.50} & 639.11 \\
Text     & 85.10 & 55.97 & 46.67 & \textbf{76.34} & 77.59 & \textbf{79.08} & \textbf{83.11} & 69.88 & 67.45 & 641.18 \\
Vision   & \textbf{88.44} & \textbf{56.51} & \textbf{47.56} & 75.73 & \textbf{77.88} & 78.72 & 82.78 & \textbf{70.52} & 68.37 & \textbf{646.50} \\
\bottomrule
\end{tabular}
}
\end{table}

\subsection{Impact of the Query Source of AGF}
A core design choice in our AGF module is the source of the query used to gate the injection of hierarchical visual features.  
Intuitively, the explicit instructions in the \textit{text tokens} seem optimal for this task. 
To test this, we ablated our standard \texttt{Visual-Query} approach against a \texttt{Text-Query} variant, where the $h_{\text{att}}$ vector in Eq.~\ref{equ:att_h} is derived solely from the text portion of the hidden state $h_{\text{t}}$.

The results, presented in Tab.~\ref{ablation_result_query}, reveal a counter-intuitive finding: the Text-Query variant, despite its intuitive appeal, is consistently outperformed by our Visual-Query approach. We attribute this superiority to two complementary mechanisms.

First, the Visual-Query approach benefits from inherent modality alignment. By using visual hidden states as the query, the fusion process functions as a residual refinement mechanism—essentially asking: "Given the current visual context, what specific fine-grained details from the incoming hierarchical features are missing?" This intra-modal comparison is significantly more direct for the attention mechanism to learn than a complex cross-modality mapping, which requires the model to first translate abstract textual semantics into the visual domain before retrieval can be performed.

Second, and perhaps more critically, relying on Text-Query introduces a premature information bottleneck. When text tokens are used to gate visual information, the AGF module effectively acts as a semantic filter that retains only the intersection of the current textual state and the incoming visual features. This ``intersection-only'' strategy risks discarding a vast amount of visual nuance that, while not explicitly requested by the current text token, may be vital for the LLM's subsequent reasoning steps.

Taken together, these two factors explain the superiority of the Visual-Query design. Crucially, because this gating and querying process occurs \textit{within} the LLM decoder—rather than as a pre-processing step—the AGF module is able to \textbf{preemptively encode} salient features from the visual hierarchy directly into the LLM's own visual token representations. This prepares a rich, continuously refined visual summary within the LLM's embedding space, which subsequent self-attention layers can efficiently cross-reference against the evolving text prompt throughout the entire decoding process.

\begin{figure*}[tp]
\includegraphics[width=1.0\textwidth]{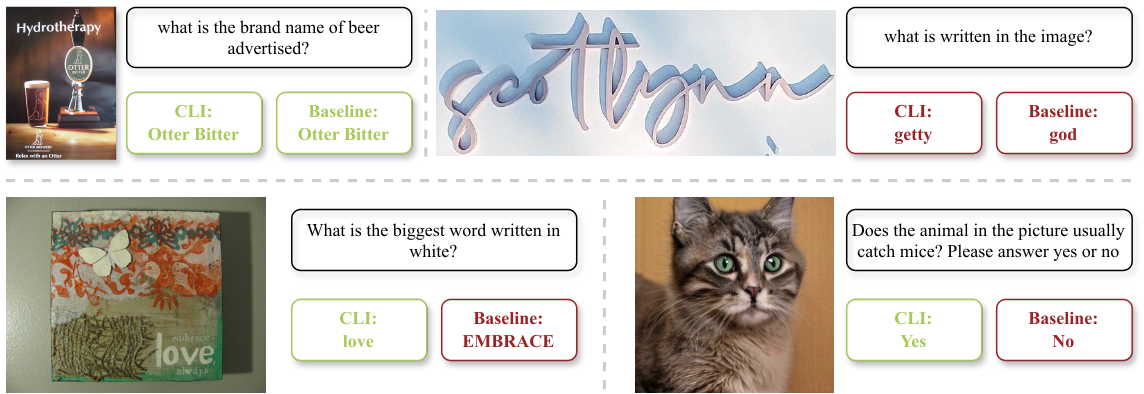}
\caption{
Qualitative comparison of CLI and the baseline on diverse visual reasoning tasks.
}
\label{fig:visualization_app}
\end{figure*}

\section{Qualitative Analysis: Strengths and Limitations}
\label{app_qualitative}
To provide an intuitive understanding of our CLI framework's performance, we present a series of qualitative case studies in Fig.~\ref{fig:visualization_app}. 

\paragraph{Complex Visual Reasoning and Instruction Following.}
The bottom-left panel provides the most compelling evidence of CLI's advanced reasoning capabilities. 
The prompt asks for the ``biggest word written in white". 
The baseline model identifies ``EMBRACE". 
This indicates a failure to grasp the holistic message of the image. 
In stark contrast, our CLI model correctly identifies ``love." 
This success demonstrates that CLI, by integrating global context with local details, enables the LLM to achieve a deeper semantic understanding of the scene. 
This showcases a leap from simple perception to sophisticated, non-literal multimodal reasoning.

\paragraph{Robustness in Challenging OCR and Hallucination Reduction.}
The top row illustrates performance on OCR tasks of varying difficulty. 
For the clear, standard font in the top-left example (``Otter Bitter"), both models perform competently, establishing a baseline. 
The top-right case, however, presents a highly stylized, artistic font that is challenging for both architectures. 
Here, the baseline model fails catastrophically, hallucinating the unrelated word ``god" due to the weak and ambiguous visual signal. 
Our CLI model, while also unable to decipher the entire word, correctly identifies the initial, more legible part, ``getty." 
This demonstrates a critical advantage of CLI: improved visual grounding. By providing richer, multi-level visual evidence, CLI anchors the LLM's output to the actual image content, effectively suppressing the tendency to invent information and thus reducing catastrophic hallucinations.

\paragraph{Reliable Common-Sense Reasoning.}
The bottom-right example probes the models' alignment with common-sense world knowledge. 
Our CLI model provides the standard, expected answer (``Yes," cats typically hunt mice). The baseline's response (``No") is unpredictable and counter-intuitive. While one could argue for a nuanced interpretation, this erratic behavior, when viewed alongside its other failures, points to a less reliable reasoning process. CLI's response, in this context, demonstrates more stable and predictable alignment with general knowledge.

In summary, these qualitative examples illustrate that CLI not only enhances fine-grained perception but, more importantly, enables the model to perform complex visual-linguistic reasoning, improves its robustness against hallucination in ambiguous scenarios, and ensures more reliable alignment with common-sense knowledge.

\end{document}